\documentclass[10pt,journal,compsoc]{IEEEtran}
\usepackage{amsmath}
\usepackage{amssymb}
\usepackage{booktabs}
\usepackage{xspace}
\usepackage{algorithm}
\usepackage{algorithmic}
\usepackage{multirow}
\usepackage{booktabs}
\usepackage{adjustbox}
\usepackage{times}
\usepackage{epsfig}
\usepackage{graphicx}
\usepackage{amsmath}
\usepackage{amssymb}
\usepackage{multirow}
\usepackage{color}
\usepackage{caption2}
\usepackage{blindtext}
\usepackage{multicol}
\usepackage{tikz}
\usepackage{amsmath}
\usepackage{booktabs}
\usepackage{adjustbox}
\usepackage{multirow}
\usepackage{epstopdf}
\usepackage{enumitem}
\usepackage{mathrsfs} 
\usepackage{enumitem}
\usepackage{mathrsfs} 
\usepackage{bm} 
\usepackage{amsfonts}
\usepackage{color}
\usepackage{xspace}
\usepackage{url}
\usepackage{multirow}
\usepackage{enumitem}
\usepackage{mathrsfs} 
\usepackage{bm} 
\usepackage{amsfonts}
\usepackage{color}
\usepackage{xspace}
\usepackage{url}
\usepackage{multirow}
\usepackage{adjustbox}
\usepackage{tikz}
\usepackage{amsmath}
\usepackage{epsfig}
\usepackage{graphicx}
\usepackage{amsmath}
\usepackage{algorithm}
\usepackage{algorithmic}
\usepackage{soul}
\usepackage{filecontents}
\usepackage{verbatim}
\usepackage{multirow}
\usepackage{threeparttable}

\usepackage{booktabs}
\usepackage{adjustbox}

\usepackage{longtable}
\ifCLASSOPTIONcompsoc
  \usepackage[nocompress]{cite}
\else
  \usepackage{cite}
\fi
\ifCLASSINFOpdf
\else
\fi

\begin{document}

\newcommand{\sys}{{\sc Megatron}\xspace}

%
\title{\sys: Evasive Clean-Label Backdoor Attacks against Vision Transformer}

\author{Xueluan Gong,
        Bowei Tian,
        Meng Xue,
        Shuike Li,
        Yanjiao Chen,~\IEEEmembership{Senior Member,~IEEE,}
        and Qian Wang,~\IEEEmembership{Fellow,~IEEE}\\
      
\thanks{X. Gong is with the School of Computer Science, Wuhan University, China. E-mail: xueluangong@whu.edu.cn.}
\thanks{M. Xue is with the Department of Computer Science and Engineering of Hong Kong University of Science and Technology, China. E-mail: csexuemeng@ust.hk}
\thanks{Y. Chen is with the College of Electrical Engineering, Zhejiang University, China. Email: chenyanjiao@zju.edu.cn.}
\thanks{B. Tian, S. Li, and Q. Wang are with the School of Cyber Science and Engineering, Wuhan University, China. E-mail:\{boweitian, lishuaike, qianwang\}@whu.edu.cn.}

}

\IEEEtitleabstractindextext{%
\begin{abstract}
Vision transformers have achieved impressive performance in various vision-related tasks, but their vulnerability to backdoor attacks is under-explored. A handful of existing works focus on dirty-label attacks with wrongly-labeled poisoned training samples, which may fail if a benign model trainer corrects the labels. In this paper, we propose \sys, an evasive clean-label backdoor attack against vision transformers, where the attacker injects the backdoor without manipulating the data-labeling process. To generate an effective trigger, we customize two loss terms based on the attention mechanism used in transformer networks, i.e., latent loss and attention diffusion loss. The latent loss aligns the last attention layer between triggered samples and clean samples of the target label. The attention diffusion loss emphasizes the attention diffusion area that encompasses the trigger. A theoretical analysis is provided to underpin the rationale behind the attention diffusion loss. 
Extensive experiments on CIFAR-10, GTSRB, CIFAR-100, and Tiny ImageNet demonstrate the effectiveness of \sys. \sys can achieve attack success rates of over 90\% even when the position of the trigger is slightly shifted during testing.
Furthermore, \sys achieves better evasiveness than baselines regarding both human visual inspection and defense strategies (i.e., DBAVT, BAVT, Beatrix, TeCo, and SAGE). 
\end{abstract}

}

\maketitle

\section{Introduction}


\IEEEPARstart{V}ision transformer (ViT) \cite{dosovitskiy2020image} is a promising alternative architecture to traditional convolutional neural networks (CNNs) for computer vision applications. By leveraging self-attention mechanisms to capture spatial relationships in images, vision transformer networks have demonstrated state-of-the-art performance across a wide range of tasks, including image classification \cite{chen2021crossvit,bhojanapalli2021understanding}, object detection \cite{beal2020toward, fang2021you}, segmentation \cite{strudel2021segmenter, gu2022multi}, autonomous driving \cite{liumonocular}, and face recognition \cite{gao2021tfe,zhong2021face}. 
These well-performed vision transformers are usually trained with a huge training dataset collected from various sources. Online public datasets, being abundant and costless, serve as an important source of training data for many model trainers. However, the quality of public datasets cannot be guaranteed. To make matters worse, public datasets may contain poisoned data samples that can enable backdoor injection through training. Backdoored models will accurately classify clean samples but deliberately misclassify any sample containing the trigger as the target label. For instance, a backdoored traffic sign recognition model may predict a ``stop sign" with a meticulously designed sticker (trigger) as the ``speed limit" (target label).

Despite the plethora of backdoor attacks targeting Convolutional Neural Networks (CNNs) \cite{chen2020backdoor, 22NDSSgong}, there is a lack of literature on backdoor attacks against vision transformers.
Different from CNNs that capture pixel-wise local features through convolutions, vision transformers extract global contextual information through image patches. Vision transformers interpolate these patches through the attention mechanism with a stronger shape recognition capacity \cite{xie2022vit}. Directly applying CNN-oriented backdoor attacks to vision transformers may result in sub-optimal attack performance. Besides, vision transformers are shown to be more robust to input perturbations than CNNs \cite{dosovitskiy2020image, shao2021adversarial}, which may be challenging for trigger design in backdoor attacks.

The majority of existing backdoor attacks against vision transformers are dirty-label attacks \cite{yuan2023you,zheng2022trojvit,lv2021dbia}, in which poisoned samples are deliberately assigned with the wrong label during training. Conspicuous fixed triggers are used for poisoning training samples, making the attacks vulnerable to human inspection and changes in trigger locations \cite{doan2022defending}. As far as we know, there is only one work on clean-label backdoor attacks against vision transformer \cite{subramanya2022backdoor}. However, Subramanya  \cite{subramanya2022backdoor} directly adapts a former CNN-oriented attack HB \cite{saha2019hidden} without considering the special characters of vision networks, resulting in sub-optimal trigger quality and attack performance.

In this paper, we propose \sys, an evasive clean-label backdoor attack customized for vision transformers. {\color{black}The crux of \sys lies in its design of covert and highly effective poisoned samples with correct labels. Rather than employing randomly chosen triggers, we present a trigger generation algorithm that meticulously incorporates two loss terms. The first loss item minimizes the distance of the last attention layer between the triggered samples and the samples of the target label. The second loss item increases the importance of the attention diffusion area of the trigger. 
Inspired by the intrinsic properties of the transformer, i.e., the input images are converted into one-dimensional vectors during the transformation process, we divide the generated trigger into sub-triggers for data poisoning, which can further improve the trigger concealment while maintaining a high attack success rate.} We have conducted extensive experiments to evaluate the attack performance of \sys. We have compared \sys with various state-of-the-art vision transformer backdoor attacks, i.e., DBIA \cite{lv2021dbia}, BAVT \cite{subramanya2022backdoor}, DBAVT-BadNets \cite{doan2022defending}, TrojViT \cite{zheng2022trojvit}, and DBAVT-WaNet \cite{nguyen2021wanet}, on four datasets, i.e., CIFAR-10, GTSRB, CIFAR-100, and Tiny ImageNet.
It is shown that \sys achieves attack success rates of up to 96\% across all datasets, exhibiting superior image quality compared to existing approaches. 
Furthermore, \sys demonstrates resilience against existing backdoor defenses, including DBAVT \cite{doan2022defending}, BAVT \cite{subramanya2022backdoor}, Beatrix \cite{ma2022beatrix}, TeCo \cite{liu2023detecting}, and SAGE \cite{23spgong}.

To conclude, we make the following key contributions.
\begin{itemize}
\item {\color{black}We develop an effective and stealthy clean-label backdoor attack framework against vision transformers. The proposed backdoor attack, \sys, can successfully inject the backdoor without accessing the data labeling process.}
\item We design two novel transformer-oriented loss terms when generating the backdoor trigger. A theoretical analysis is provided to clarify the contribution of attention diffusion loss to enhancing the attack performance. We also propose to use sub-triggers through masking operations to significantly enhance trigger concealment while maintaining the attack success rate.
\item Extensive experiments conducted on CIFAR-10, GTSRB, CIFAR-100, and Tiny ImageNet showcase the effectiveness and evasiveness of \sys. \sys can evade human visual inspection and existing defense strategies.

\end{itemize}

\section{Preliminaries}\label{Background}

\subsection{Vision Transformer}

The transformer architecture was initially designed for natural language processing (NLP) \cite{vaswani2017attention} tasks. 
Unlike convolutional neural networks, the transformer network relies on the attention mechanism to process sequences of input tokens in parallel.

Recently, the transformer network has been adapted to computer vision tasks, modeling relationships between different parts of an image using the self-attention mechanism. A popular vision transformer is ViT \cite{dosovitskiy2020image}.
Let $\mathbf{x} = \{\mathbf{x}_1, \mathbf{x}_2, ..., \mathbf{x}_n\}$ be an image sample consisting of a sequence of $n$ patches. Each patch is represented as a $p \times p \times c$-dimensional tensor, with $p \times p$ pixels and $c$ channels. ViT first applies an embedding layer to each patch to convert it into a $d$-dimensional embedding vector $\mathbf{e}_i = \mathcal{E}(\mathbf{x}_i)$.
Next, ViT applies a series of transformer encoder layers to the embedding vectors. Each encoder layer comprises two sub-layers, i.e., a multi-head self-attention mechanism (MHSA) layer and a position-wise feed-forward network (FFN) layer. The MHSA layer models interactions between embedding vectors using self-attention, and the FFN layer applies a non-linear transformation to each patch embedding. 
The attention mechanism in the MHSA layer can be decomposed into two main operations, i.e., attention rollout and attention diffusion. The attention rollout operation computes the similarity between each query vector and all key vectors using the dot product, scales the dot products by $\sqrt{d}$ to avoid gradient explosion, applies a softmax function to obtain attention weights, and finally computes a weighted sum of the value vectors. The attention rollout can be expressed as 
\begin{equation}
\label{equ:attention}
\mathcal{A}(Q,K,V) = \text{softmax}\left(\frac{QK^T}{\sqrt{d}}\right)V,
\end{equation}
where $Q$, $K$, and $V$ are the query, key, and value matrices, respectively, and $d$ is the dimensionality of the key vectors.

The attention diffusion operation computes multiple attention heads in parallel and concatenates the resulting vectors along the last dimension. The concatenated vectors are then linearly transformed to obtain the final output. The attention diffusion operation can be expressed as
\begin{equation}
\begin{split}
\mathcal{M}(Q,K,V) &= \text{Concat}\left(\text{head}_1,\ldots,\text{head}_h\right)W^O,\\
\text{head}_i &= \mathcal{A}(QW_i^Q,KW_i^K,VW_i^V),
\end{split}
\end{equation}
where $h$ is the number of attention heads, $W_i^Q$, $W_i^K$, and $W_i^V$ are learnable weight matrices for the $i$-th attention head, and $W^O$ is a learnable weight matrix that maps the concatenated output of all heads to the output. 
\begin{figure*}[htbp]
    \centering
    \scriptsize
	\includegraphics[trim=0mm 0mm 0mm 0mm, clip,width=7in]{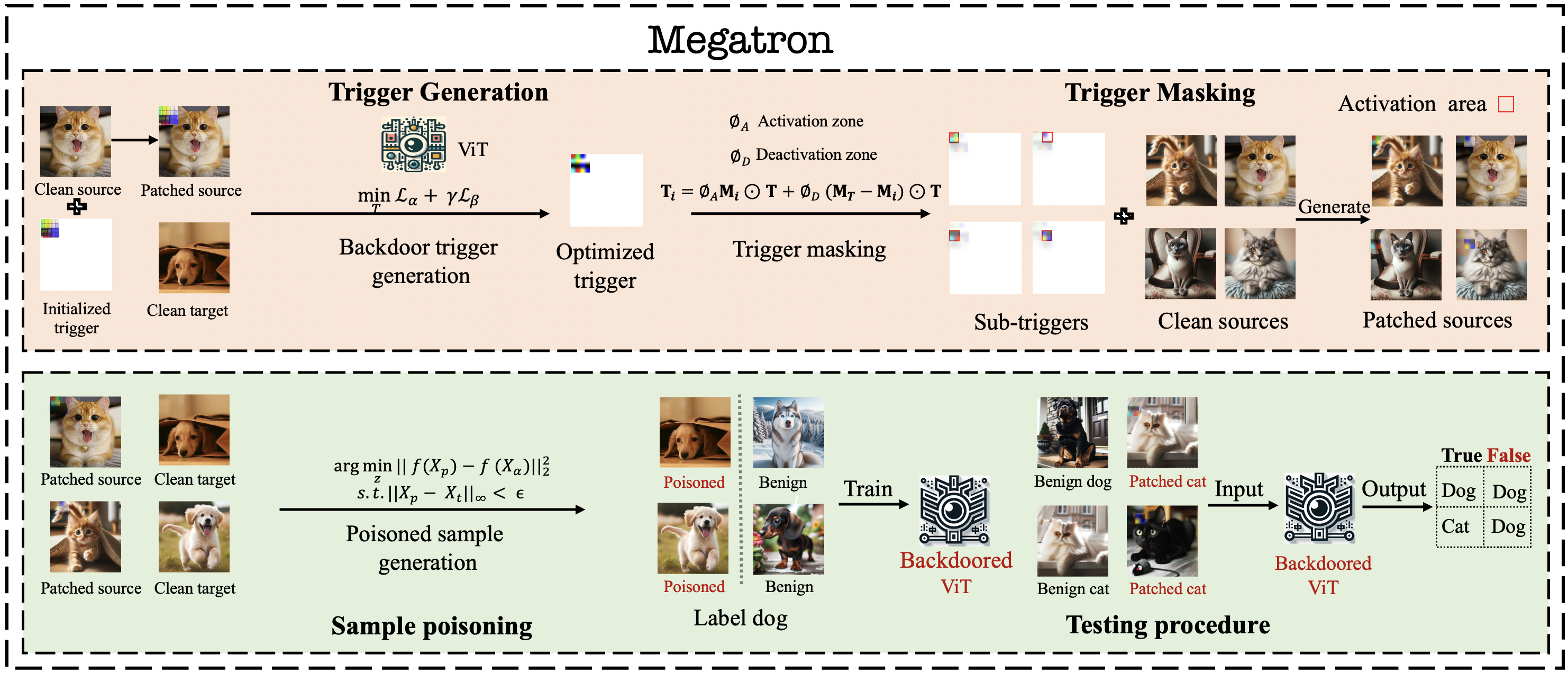}\\
	
	\caption{Overview of \sys. \sys features three key components, i.e., trigger generation, trigger masking, and sample poisoning. }
	\label{fig:overview}
\end{figure*}

\subsection{Backdoor Attack}
Backdoor attacks manipulate the training process of a model such that the backdoored model will produce a specific or wrong output when presented with a triggered input \cite{chen2020backdoor,liu2022piccolo}. To achieve stealthiness, the trigger should be imperceptible to humans, and the main task of the model should be unaffected, i.e., the model accuracy on clean inputs should be high.
\begin{equation}
f_b(\mathbf{x}) = \begin{cases}
t, & \text{if } \mathbf{x} \text{ contains a trigger}, \\
f_c(\mathbf{x}), & \text{otherwise},
\end{cases}
\end{equation}
where $\mathbf{x}$ is an input sample, $f_b$ is a backdoored model, $f_c$ is a clean model, and $t$ is the target false label.

\subsubsection{Backdoor Attacks against CNN} 
Early works on backdoor attacks mainly focused on Convolutional Neural Networks (CNNs) \cite{pang2020tale}.
The trigger plays a crucial role in backdoor attacks since it interacts with the backdoored model to cause misclassification. Various trigger-generation methods have been proposed in the literature for backdoor attacks.
Gu et al. \cite{gu2019badnets} proposed the first backdoor attack, i.e., BadNets, which randomly designated a simple pattern, such as a square with random pixel values, as the trigger. 
Salem et al. \cite{salem2020dynamic} proposed a dynamic trigger generation algorithm based on a generative network. To enhance the attack robustness, Li et al. \cite{li2020rethinking} varied the trigger location and appearance. Besides, Liu et al. \cite{liu2017trojaning} designed a model-dependent trigger generation algorithm that establishes a strong connection between the trigger-excited neuron and the output. Gong et al. \cite{Gong2021} improved the model-dependent trigger generation approach by enhancing the neuron selection criterion. 

Visible triggers in many backdoor attacks \cite{gu2019badnets,liu2017trojaning,salem2020dynamic,nguyen2020input,Gong2021,ji2017backdoor,chen2017targeted,wang2019neural} are easy to detect during both training and inference phases. Therefore, recent research has been dedicated to hiding the trigger. For example, Liao et al. \cite{liao2018backdoor} proposed to use the pixel differences between the original and adversarial samples as the backdoor trigger. Li et al. \cite{li2020invisible} formulated the trigger generation process as a bilevel optimization problem and optimized the trigger to boost a group of neuron activations through L$_p$-regularization to achieve invisibility. Gong et al. \cite{22NDSSgong} introduced a quality of experience (QoE) term into the loss function and adjusted the trigger transparency value to hide the backdoor trigger. Saha et al. \cite{saha2019hidden} proposed HB, which makes poisoned samples share a similar pixel space with benign samples of the target label. Nguyen et al. \cite{nguyen2021wanet} proposed to use a small and smooth warping field to generate the backdoored images to achieve the invisibility goal.

\subsubsection{Backdoor Attack against Vision Transformer}

To the best of our knowledge, backdoor attacks against vision transformers are largely under-explored, with only a few existing works so far. Lv et al. \cite{lv2021dbia} proposed DBIA, which leveraged the attention mechanism of transformers to generate triggers and injected the backdoor using data poisoning. Zheng et al. \cite{zheng2022trojvit} proposed TrojViT, which generated the trigger using patch salience ranking and injected the backdoor as vulnerable bits in model parameters using attention-target loss. However, the triggers of DBIA and TrojViT are visible to human eyes. Recently, Yuan et al. \cite{yuan2023you} proposed BadViT, which generated the trigger based on the self-attention mechanism and adopted the blending strategy \cite{chen2017targeted} to achieve trigger invisibility. However, DBIA, TrojViT, BadViT are dirty-label attacks, which may not be successful if the attacker is a data provider. A data provider can only provide the poisoned data but cannot control the training process. Since the labels of all poisoned samples in dirty-label attacks are incorrect, the benign model trainer may re-label poisoned samples and defuse such attacks. Existing backdoor attacks against vision transformer largely re-use CNN-oriented backdoor attack methods, such as BAVT \cite{subramanya2022backdoor} (based on HB \cite{saha2019hidden}) and DBAVT \cite{doan2022defending} (based on WaNet \cite{nguyen2021wanet}). Different from CNNs that capture pixel-wise local features through convolutions, vision transformers extract global contextual information through patches and attentions. Applying CNN-oriented backdoor attacks directly to vision transformers yields a relatively low attack success rate \cite{subramanya2022backdoor}.

To defend against backdoor attacks on vision transformers, Subramanya et al. \cite{subramanya2022backdoor} proposed a test-time backdoor defense that relies on the interpretation map. Doan et al. \cite{doan2022defending} introduced a defense that mitigates backdoor attacks using patch processing. The intuition of these defenses is that clean data accuracy and backdoor attack success rates of vision transformers respond differently to patch transformations before the positional encoding.

In this paper, we present a novel clean-label backdoor attack against vision transformers. \sys outperforms existing attacks by demonstrating superior trigger imperceptibility and a higher attack success rate. Furthermore, it is shown that \sys is also effective in evading state-of-the-art backdoor defenses.

\subsection{Threat Model}

In \sys, we assume a more pragmatic scenario where the attacker is merely a data provider. The attacker does not have access to the training process of the vision transformer. The attacker can only provide poisoned samples that are collected as training data samples for a vision transformer. The attack goal is to craft poisoned samples that deceive the victim into training a backdoored vision transformer with high prediction accuracy on clean input but targeted false prediction on triggered inputs. The poisoned samples must be natural enough to evade human visual inspections. 

We set the following limitations for the attacker.
\begin{itemize}
	
      \item \emph{No control over data labeling}. The attacker has no access to the data labeling process. Therefore, a clean-label attack is required to ensure correct labels on poisoned samples.  
      
      \item \emph{No control over model training}. The attacker does not know the model architecture and training methodology chosen by the model trainer.

      \item \emph{No knowledge of potential defense strategies}. The attacker does not know whether or not defense strategies are employed by the model trainer. If the model trainer indeed adopts defense strategies, the attacker does not know the technical details of those defense strategies.

\end{itemize}

\begin{algorithm}[tt]
	\caption{Algorithm of Poisoned Sample Construction} \label{alg:trigger injection}
	\begin{algorithmic}[1]
		\REQUIRE Local surrogate vision transformer $f$, initial trigger $\mathbf{T_0}$, trigger $\mathbf{T}$, training data dataset $D$, 
        threshold $\tau$, the maximum number of iterations of trigger generation $E$, the maximum number of iterations of sample poisoning $E'$, and learning rate $lr$.

		\ENSURE The poisoned samples $\mathbf{x}_p$.
		\STATE $Loss = INF, memset(T_0, 0)$
        \WHILE {$Loss > \tau$ \textbf{and} $i < E$}
          \FORALL{$x \in D$}
            \STATE $\mathcal{L}_{\alpha}=|| \mathcal{A}_N(\mathbf{x}_{t})-\mathcal{A}_N(\mathbf{x}_c)||_2^2$
            \STATE $\mathcal{L}_{\beta}= \sum_{i \in \mathcal{D}}(1-\mathbf{A}_N^{0,i})+\sum_{i \notin \mathcal{D}} \mathbf{A}_N^{0,i}$
            \STATE // Apply gradient surgery method
		  \STATE $\delta=PCGrad(\mathcal{L}_{\alpha}, \gamma \mathcal{L}_{\beta})$
            \STATE $\mathbf{T_0} = \mathbf{T_0} - lr \cdot \delta$
            \STATE $i = i+1$
         \ENDFOR
        \ENDWHILE
        \STATE $\mathbf{T}$ = $\mathbf{T_0}, i = 0$
        \STATE $\mathbf{T}_i = \phi_A  \mathbf{M}_i \odot \mathbf{T} + \phi_D  (\mathbf{M}_T-\mathbf{M}_i) \odot \mathbf{T}$
        \WHILE {$\mathcal{L} > \tau$ \textbf{and} $i < E'$}
        \FORALL{$x \in D$}
        \STATE $\mathcal{L} = || f(\mathbf{x}_p)-f(\mathbf{x}_a)||_2^2$
        \STATE {\color{black}$\quad||\mathbf{x}_p - \mathbf{x}_t||_\infty < \epsilon \nonumber$}
        \STATE $\delta = \frac{\partial{\mathcal{L}}}{\partial \mathbf{x}_p}$
        \STATE $\mathbf{x}_p = \mathbf{x}_p - lr\cdot \delta$
        \STATE $i = i+1$
        \ENDFOR
        \ENDWHILE

        \RETURN Poisoned samples $\mathbf{x}_p$.
  \end{algorithmic}
\end{algorithm}

\section{\sys: Detailed Construction}

The \sys framework comprises three pivotal stages, i.e., trigger generation, trigger masking, and sample poisoning. The trigger generation stage focuses on crafting effective triggers for the backdoor attack. The trigger masking stage produces sub-triggers to enhance the attack in terms of stealthiness. Finally, the sample poisoning stage constructs poisoned samples based on the generated trigger. Fig.~\ref{fig:overview} illustrates the overall procedure of \sys. Algorithm~\ref{alg:trigger injection} outlines the entire backdoor attack process of \sys.

\subsection{Trigger Generation}

Backdoor attacks that employ random triggers can be easily mitigated by current defenses \cite{li2021invisible,li2020rethinking}. To enhance attack efficacy, rather than using a random trigger, we design a novel trigger optimization algorithm.

Given a pre-determined trigger shape (i.e., rectangle) and size, we first initialize a trigger by randomly sampling from a uniform distribution $\mathcal{U}(0,1)$. Subsequently, we design two loss terms to optimize the trigger. The first loss is the latent loss $\mathcal{L}_{\alpha}$, which aims to minimize the distance between the patched source (clean source image patched with the trigger) and the sample of the target label regarding the last layer's attention. We choose the last layer since it aggregates the representations formed by the network over all of its previous layers \cite{abnar2020quantifying}. It is at this juncture that the model has the opportunity to integrate information from all parts of the image across all patches, which is vital for a comprehensive contextual understanding and accurate predictions. By utilizing solely the last layer, we streamline the computational demands.
\begin{equation}
    \mathcal{L}_{\alpha}=|| \mathcal{A}_N(\mathbf{x}_{t})-\mathcal{A}_N(\mathbf{x}_c)||_2^2,
\end{equation}
where $N$ is the last layer of the transformer, $\mathbf{x}_t$ is a sample of the target label, and $\mathbf{x}_c$ is a clean source image patched with the trigger. The latent loss encourages the output of patched samples to get close to the output of target label samples. 

The second loss is attention diffusion loss $\mathcal{L}_{\beta}$. 
Our intuition is to increase the importance of the attention diffusion area that encompasses the trigger and decrease the importance of the non-diffusion area during training. 
Formally, $\mathcal{L}_{\beta}$ is defined as
\begin{equation}
\begin{split}
    \mathcal{L}_{\beta}= \sum_{i \in \mathcal{D}}(1-\mathbf{A}_N^{0,i})+\sum_{i \notin \mathcal{D}} \mathbf{A}_N^{0,i},
\end{split}
\end{equation}
where $\mathcal{D}$ represents the attention diffusion area of the transformer. $\mathbf{A}_N$ represents the input grad-attention rollout of the last layer, which measures the importance of each token in the final prediction. Note that $\mathbf{A}_N$ is a matrix, and $\mathbf{A}_N^{0,i}$ is the item at the 0-th row and $i$-th column. 
\begin{equation}
    \mathbf{A}_l=\begin{cases}
  \mathcal{A}_l\frac{\partial y_{t}}{ \partial \mathcal{A}_l} \mathbf{A}_{l-1}, & \text{if } l > 0,
  \\
 \mathcal{A}_l\frac{\partial y_{t}}{ \partial \mathcal{A}_l}, & \text{if }  l = 0,
  \end{cases}
\end{equation}
where $y_{t}$ represents the output of the target label, and $\mathcal{A}_l$ is the attention weight of the $l$-th layer $(l=1,2, \cdots, N)$ of the transformer, computed according to Equation \eqref{equ:attention}. 
Note that the token is the unit of the input sequence in a vision transformer \cite{abnar2020quantifying}. In our experiments, we resize samples to $224 \times 224$, each sample containing $14 \times 14$ tokens. 
The attention diffusion area is equal to or larger than the trigger. 
In general, a larger diffusion area enhances the influence of the trigger, but an excessively large diffusion area may deteriorate the attack performance.
We theoretically explain the effectiveness of $\mathcal{L}_{\beta}$ on the attack performance in Section~\ref{8}. 
Thus, the overall trigger generation loss of \sys is defined as 
\begin{equation} \label{equ:trigger}
   \min_\mathbf{T} \mathcal{L}_{\alpha}+ \gamma \mathcal{L}_{\beta},
\end{equation}
where $\gamma$ is the hyperparameter that balances the two loss terms, and $\mathbf{T}$ is the trigger.

Note that the attacker utilizes a local surrogate vision transformer $f$ to compute the attention for each layer and the gradient attention rollout since the attacker cannot access the victim vision transformer model. Nevertheless, \sys demonstrates a high attack success rate even when the victim and the attacker employ different vision transformer structures.

To prevent gradient conflicts in gradient descent, we apply the projecting conflicting gradients (PCGrad) \cite{yu2020gradient} method to minimize the loss function. Specifically, PCGrad projects gradients of $\mathcal{L}_{\alpha}$ and $\mathcal{L}_{\beta}$ onto a shared subspace. This subspace minimizes conflicts by maximizing the angle between the two gradients, ensuring they do not interfere with one another. The gradient with PCGrad is defined as
\begin{equation}
\begin{cases}
\Delta\mathcal{L}_{\alpha} + \Delta\mathcal{L}_{\beta}, &\text{if} \cos(\mathcal{L}_{\alpha}, \mathcal{L}_{\beta}) > 0, \\
\Delta\mathcal{L}_{\alpha} + \frac{\Delta\mathcal{L}_{\alpha}\cdot \Delta\mathcal{L}_{\beta}}{||\Delta\mathcal{L}_{\beta}||^2} \cdot \Delta\mathcal{L}_{\beta}, & \text{otherwise},
\end{cases}
\end{equation}
where $\Delta \mathcal{L}_{\alpha} = \partial \mathcal{L}_{\alpha} / \partial \mathbf{T}, \Delta \mathcal{L}_{\beta} = \partial \mathcal{L}_{\beta}/ \partial \mathbf{T}$, and $\cos(\cdot, \cdot)$ is the cosine similarity function.

\begin{figure*}[htbp]
    \centering
    \scriptsize
	\includegraphics[trim=0mm 0mm 0mm 0mm, clip,width=7in]{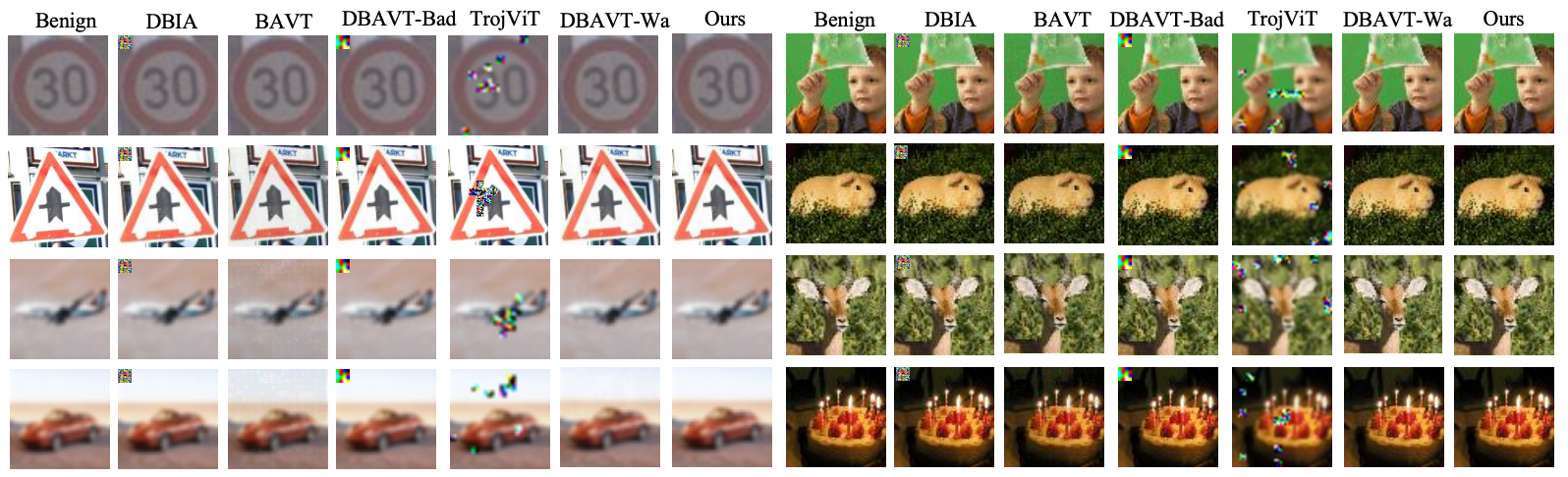}\\
	
	\caption{Compare backdoored samples generated by baseline attacks and \sys. DBAVT-Bad is DBAVT-BadNet, and DBAVT-Wa is DBAVT-WaNet.}
	\label{fig:com}
 \vspace{-0.4cm}
\end{figure*}
\subsection{Trigger Masking}
Rather than using the entire trigger $\mathbf{T}$ optimized by Equation \eqref{equ:trigger}, we propose to split the trigger into multiple sub-triggers for sample poisoning. The experimental results show that our trigger split approach significantly improves trigger stealthiness while maintaining a high attack success rate. 
This improvement can be attributed to the intrinsic properties of the transformer network, as the input images are converted into one-dimensional vectors during the transformation process. Additionally, transformers exhibit relatively low variance in feature maps \cite{xie2022vit}, enabling a more adaptable trigger form than CNN models. 

Specifically, we split the original trigger into $K$ areas with mask $\mathbf{M}_i$ corresponding to each area. Sub-trigger $\mathbf{T}_i$ is formed as:
\begin{equation}
\mathbf{T}_i = \phi_A  \mathbf{M}_i \odot \mathbf{T} + \phi_D  (\mathbf{M}_T-\mathbf{M}_i) \odot \mathbf{T},
\end{equation}
where $\odot$ denotes element-wise multiplication, and $\mathbf{M}_T$ is the entire trigger mask. $\phi_A$ and $\phi_D$ are the transparency values for the masked and unmasked areas. 

The lower the transparency value is, the more imperceptible the trigger is. Our user study shows that when the transparency values of $\phi_A$ and $\phi_D$ are set as 0.5 and 0.1, the human eyes can hardly discern the trigger. Thus, we set $\phi_A=0.5$ and $\phi_D=0.1$ by default. We also evaluate the impact of the transparency value on the attack performance of \sys in evaluations. Note that during the attack, the adversary is able to use any sub-trigger $T_i$ to activate the backdoor and induce misclassification. The availability of multiple small-sized sub-triggers makes \sys more difficult to be defended against, as verified by our evaluations.

\subsection{Sample Poisoning}
In dirty-label backdoor attacks, poisoned samples are incorrectly labeled as the target label, making it relatively easy for defenders to detect and remove poisoned samples.
To improve attack stealthiness, we materialize \sys as a clean-label backdoor attack in which poisoned samples are labeled correctly. To enable clean-label attacks, we construct poisoned samples that resemble target label samples in the pixel space but resemble the patched source in the feature space.

To create triggered samples, we randomly select $Q$ clean samples, divided into $\lceil\frac{Q}{K} \rceil$ sets, where $K$ is the number of available sub-triggers. Given the $i$-th sample in the $k$-th set and the sub-trigger $T_i$, we impose the trigger on to generate a triggered sample $\mathbf{x}_{i,k}+\mathbf{T}_i$. 

Inspired by HB \cite{saha2019hidden}, we optimize the feature space of the poisoned sample to closely match the patched source and the pixel space of the poisoned sample to closely resemble the sample from the target label.
\begin{align}
\label{attention distance}
     &\min\limits_{\mathbf{x}_p}|| f(\mathbf{x}_p)-f(\mathbf{x}_a)||_2^2, \\
    & \text{s.t., } ||\mathbf{x}_p - \mathbf{x}_t||_\infty < \epsilon \nonumber,
\end{align}
where $f(\cdot)$ represents the feature extracted by the transformer model, $\mathbf{x}_p$ is the poisoned sample, $\mathbf{x}_a$ is a patched source, $\mathbf{x}_t$ is a sample of the target label. $\epsilon$ is a minor value to guarantee that the poisoned image $\mathbf{x}_p$ remains visually indistinguishable from the target image $\mathbf{x}_t$.

\section{Experiment Setup}
\begin{table*}[tt]
	\caption{Comparison of \sys with DBIA \cite{lv2021dbia}, BAVT \cite{subramanya2022backdoor}, DBAVT-BadNets \cite{doan2022defending}, TrojViT \cite{zheng2022trojvit}, and DBAVT-WaNet \cite{nguyen2021wanet}.} 
	\label{tab:com}
	\centering
	\footnotesize
 \setlength\tabcolsep{7pt}
	\begin{tabular}{lllccccccccccc}
		\toprule
		&&CDA
           &SCDA&SASR &PSNR&SSIM&LPIPS&L$_1$ distance \\
			\cline{3-9}
		\multirow{6}{*}{\shortstack{CIFAR-10}}&DBIA&97.46\%&97.85\%&	 98.02\%&9.498306277&0.97032434&	0.111538727	&0.045788546 \\
            &BAVT&96.40\%&84.52\%&80.33\%&24.78704153&0.96168715&	0.084357165&0.026071861\\
            &DBAVT-BadNets&96.00\%&100.0\%&100.0\%&24.02675079&0.96076837&0.084862622&0.030870578\\
            &TrojViT&96.98\%&97.00\%&97.80\%&5.172618398&	0.90867385&0.599903807&0.086292282\\
            &DBAVT-WaNet&83.19\%&83.98\%&19.67\%&	42.55067178&0.99651533&0.002949054&0.004764633\\
            &\sys&96.38\%&91.06\%&96.03\%&40.95445537&0.98685187&0.020913042&0.024609151\\
		\midrule
		&&CDA
            & SCDA&SASR&PSNR&SSIM&LPIPS&L$_1$ distance \\
		\cline{3-9}
		\multirow{6}{*}{\shortstack{GTSRB}}&DBIA&96.56\%&95.64\%&96.05\%&9.229372641&0.97033067 & 0.109800851&	0.047954460 \\
            &BAVT&96.45\%&86.82\%&82.15\%&23.98223772&	0.96972583&0.068876714&0.027542898\\
            &DBAVT-BadNets&88.00\%&94.07\%&92.20\%&23.51824899&	0.96956628&0.061666282&0.032136891\\
            &TrojViT&99.09\%&94.00\%&96.03\%&4.47244447&0.91404196&	0.539841508&0.089894254\\
            &DBAVT-WaNet&81.75\%&82.09\%&4.64\%&	44.17835503&0.99707979&0.002103633&0.003752139	\\
            &\sys&95.06\%&92.01\%&96.25\%&42.16052373&
            0.98821563&0.014863365&0.021438070\\
           \midrule
		&&CDA
            & SCDA&SASR&PSNR&SSIM&LPIPS&L$_1$ distance \\
		\cline{3-9}
		\multirow{6}{*}{\shortstack{CIFAR-100}}&DBIA&90.20\%&89.87\%&98.00\%&9.45239590&0.97039095&0.114754133&0.045978279 \\
            &BAVT&88.40\%&82.26\%&78.38\%&23.32390113&0.96604105&0.071154475&	0.029951381\\
            &DBAVT-BadNets&94.00\%&92.06\%&99.08\%&
            22.99643840&0.96068738&0.065457862&0.032614115\\
            &TrojViT&87.84\%&90.03\%&96.00\%&5.40328789&0.90887051&0.537958621&0.080050066\\
            &DBAVT-WaNet&63.42\%&64.11\%&9.82\%&49.13612167&0.99918764&0.000630564&0.002330421\\
            &\sys&87.28\%&92.03\%&96.19\%&    41.22737319&0.98680115&0.017893705&0.025383241\\
            \midrule
		&&CDA
            & SCDA&SASR&PSNR&SSIM&LPIPS&L$_1$ distance \\
		\cline{3-9}
		\multirow{6}{*}{\shortstack{Tiny ImageNet}}&DBIA&81.25\%&82.41\%&100.0\%&9.38644366&0.97084104&0.087757164&0.046567879\\
            &BAVT&92.40\%&84.80\%&87.22\%&24.29126279&	0.97306752&0.067220901&0.026612388\\
            &DBAVT-BadNets&88.45\%&96.04\%&96.02\%&23.93382611&	0.97255355&0.071769386&0.028993654\\
            &TrojViT&81.18\%&90.00\%&96.00\%&6.44871437&0.91046688&
            0.544488658&0.071984994\\
            &DBAVT-WaNet&77.47\%&83.69\%&15.29\%&	30.65163385&0.97139597&0.008564807&	0.016684100\\
            &\sys&88.03\%&91.10\%&96.36\%&42.22344340&0.98954761&	0.016416840&	0.012017685\\
            \bottomrule
	\end{tabular}
	
\end{table*}
\subsection{Datasets and Models}
We conduct experiments on various vision tasks, covering multiple datasets, including CIFAR-10 \cite{krizhevsky2009learning}, GTSRB \cite{StallkampSSI12}, CIFAR-100 \cite{krizhevsky2009learning}, and Tiny ImageNet \cite{le2015tiny}. {\color{black}We trained ViT-base models \cite{dosovitskiy2020image} on these datasets as local surrogate ViT models. The default model structure for the victim is the DeiT \cite{touvron2021training} model. We also have assessed the effectiveness of \sys when the victim chooses other vision transformer structures.}

\textbf{CIFAR-10}: CIFAR-10 \cite{krizhevsky2009learning} consists of 60,000 images belonging to 10 classes, with each class containing 6,000 images. 
We randomly selected 50,000 samples for the training set and the remaining 10,000 samples for the test set. We trained a local surrogate vision transformer on the training set for 7 epochs, using a learning rate of 0.01 and a batch size of 32. The trained ViT model can achieve a prediction accuracy of 97.72\% on the test set.

\textbf{GTSRB}: GTSRB \cite{StallkampSSI12} contains images of German traffic signs that belong to 43 classes. The dataset is divided into 39,209 training samples and 12,630 testing samples. 
We trained a local surrogate vision transformer on the training set for 10 epochs, using a learning rate of 0.01 and a batch size of 32. The trained ViT model can achieve a prediction accuracy of 96.77\% on the test set.

\textbf{CIFAR-100}: CIFAR-100 \cite{krizhevsky2009learning} includes 600,000 images that belong to 100 classes, with each class containing 500 training samples and 100 testing images. 
We trained a local surrogate vision transformer on the training set for 10 epochs, using a learning rate of 0.01 and a batch size of 32. The trained ViT model can achieve a prediction accuracy of 88.45\% on the test set.

\textbf{Tiny ImageNet}: ImageNet \cite{krizhevsky2017imagenet} is a large-scale image recognition dataset that contains more than 14 million labeled images across 22,000 categories. Due to the limited computing source, we adopt a subset of ImageNet, i.e., Tiny ImageNet \cite{le2015tiny} to conduct the experiments. Tiny ImageNet has 200 classes, each with 500 training images, 50 validation images, and 50 test images.
We trained a local surrogate vision transformer on the training set for 10 epochs, using a learning rate of 0.01 and a batch size of 32. The trained model can achieve a prediction accuracy of 88.18\% on the test set.

\subsection{Evaluation Metrics}

As a clean-label backdoor attack, following HB \cite{saha2019hidden}, in this paper, we mainly focus on one-to-one attacks. The goal is to make the backdoored model misclassify trigger-imposed samples of a specific source label to the target label. Note that we also assess and demonstrate the efficacy of \sys under any-to-one attacks. In this setting, the objective is to make the backdoored model misclassify trigger-imposed samples of any labels to the target label.
We evaluate the effectiveness and evasiveness of \sys with 7 evaluation metrics, including clean data accuracy (CDA), source attack success rate (SASR), source clean data accuracy (SCDA), structural similarity index measure (SSIM), peak signal-to-noise ratio (PSNR), learned perceptual image patch similarity (LPIPS), and L$_1$ distance. CDA, SASR, and SCDA measure the effectiveness of \sys. SSIM, PSNR, LPIPS, and L$_1$ distance measure the stealthiness of the trigger.

\textbf{CDA}. CDA measures whether the backdoored model can maintain high prediction accuracy of clean data samples. 
\begin{equation}
CDA(f_b, \mathcal{X})= \frac{1}{|\mathcal{X}|} \sum_{{x}\in \mathcal{X}} \mathbf I_{[f_b({x})=y]},
\end{equation}
where $x$ is a sample of the clean dataset $\mathcal{X}$, and $y$ is the ground-truth label of ${x}$.

\textbf{SASR}. In one-to-one attacks, SASR is computed as the probability that trigger-imposed samples of the source label are misclassified to the target label.
\begin{equation}
ASR(f_b,\mathcal{X_S})=\frac{1}{|\mathcal{X_S}|} \sum_{x\in \mathcal{X_S}} \mathbf I_{[f_b(x+T_i)=t]},
\end{equation}
where $x$ is the clean sample of the source label dataset $\mathcal{X_S}$.

\begin{table*}[tt]
	\caption{Comparison of \sys with DBIA \cite{lv2021dbia}, BAVT \cite{subramanya2022backdoor}, DBAVT-BadNets \cite{doan2022defending}, TrojViT \cite{zheng2022trojvit}, and DBAVT-WaNet \cite{nguyen2021wanet} in any-to-one attacks. }
	\label{tab:com2}
	\centering
 \footnotesize
        \setlength\tabcolsep{8pt}
	\begin{tabular}{l|cccccc}
		\toprule
		\multirow{2}{*}{\shortstack{Method}}&\multicolumn{6}{c}{\color{black}CIFAR-10}  \\
             & CDA&ASR$^{\rm *}$&PSNR&SSIM&LPIPS&L$_1$ distance \\
		\hline
		DBIA&96.32\%&	98.40\%&9.498306277&0.97032434&	0.11153872	&0.045788546\\
            BAVT&94.40\%&87.20\%&24.29126279&0.96984532&0.06722090&	0.026612388\\
            DBAVT-BadNets&96.00\%&98.10\%&23.71782612& 	0.96896010 &0.08735909 &0.031357202 \\
            TrojViT&98.96\%&99.74\%&5.14141308&0.91241727&0.61245163&0.086870462\\
            DBAVT-WaNet&97.57\%&10.07\%&48.83712140&0.99865401&0.00111286&0.002102883\\
            \sys&97.15\%&98.92\%&35.68797644&0.96721107&0.08209506&0.079858869\\
		\midrule
		\multirow{2}{*}{\shortstack{Method}}&\multicolumn{6}{c}{\color{black}GTSRB}\\
            & CDA&ASR&PSNR&SSIM&LPIPS&L$_1$ distance \\
		\hline
		DBIA&96.07\%&	96.80\%&9.229372641&0.97033067 & 0.10980085&	0.047954460\\
            {BAVT}&86.45\%&78.03\%&23.45269425&0.97012485&0.06783522&0.031815644\\
            DBAVT-BadNets&88.00\%&97.50\%&22.58075594&0.96958321&0.08057846&	0.035124150	\\
            TrojViT&99.48\%&98.96\%&	5.39314620&0.91888680&0.53404443&0.084501677\\
            DBAVT-WaNet&99.13\%&4.50\%&	50.75901875&0.99881482&0.00053918&0.002055686\\
            \sys&92.09\%&94.41\%&39.94159666&0.98472887&0.028565167&	0.031615723\\
            \midrule
		\multirow{2}{*}{\shortstack{Method}}&\multicolumn{6}{c}{\color{black}CIFAR-100}\\
            & CDA&ASR&PSNR&SSIM&LPIPS&L$_1$ distance \\
		\hline
	 
  DBIA&90.33\%&97.60\%&9.45239590&0.97039095&0.11475413&0.045978279 \\
            {BAVT}&84.40\%&81.30\%&23.31849251&0.96521573&0.07105962&0.034516015\\
            DBAVT-BadNets&90.00\%&98.20\%&22.34716330& 	0.97058266&0.07006657&0.035069255 \\
            TrojViT&88.02\%&98.96\%&5.86652204&0.91242411&0.57811076&0.079721999\\
            DBAVT-WaNet&87.37\%&1.67\%&	48.27763831&0.99844939&0.00117087&0.002333022\\
            \sys&84.74\%&99.40\%&35.43962310&0.96642988&0.09823335&	0.083038695	\\
            \midrule
		\multirow{2}{*}{\shortstack{Method}}&\multicolumn{6}{c}{\color{black}Tiny ImageNet}\\
            & CDA&ASR&PSNR&SSIM&LPIPS&L$_1$ distance \\
		\hline
		DBIA&81.25\%&100.0\%&9.38644366&0.97084104&0.08775716&0.046567879 \\
            BAVT&78.84\%&83.98\%&24.12858651&0.97012561&0.06638167&	0.028590012\\
            DBAVT-BadNets&88.45\%&96.00\%&23.93382611& 	0.97255355 &0.07176938 &0.027813941 \\
            TrojViT&91.93\%&96.00\%&	6.41666838&0.91378632&0.51033226&0.072979480\\
            DBAVT-WaNet&80.22\%&69.25\%&	40.19697062&0.99209707&	0.00585494&0.005845252\\
            \sys&90.00\%&98.00\%&40.86731687&0.99055898&	0.01513319&0.025638208\\
            \bottomrule
	\end{tabular}
 \label{tab:Defenses}

 \begin{tablenotes}
\item  {\footnotesize $^*$ In any-to-one scenario, we switch from using the metric SASR to ASR, which assesses the success of backdoor attacks by calculating the probability that any trigger-imposed sample is misclassified to the target label.} 
 \end{tablenotes}
\end{table*}

\begin{table*}[tt]
	\caption{Impact of different loss functions on trigger generation.}
	\label{tab:ablation-trigger generation}
	\centering
	\setlength\tabcolsep{7pt}
	\footnotesize
	\begin{tabular}{llccccccccc}
		\toprule
		&&
            CDA & SCDA&SASR&PSNR&SSIM&LPIPS&L$_1$ distance \\
			\cline{3-9}
		\multirow{3}{*}{\shortstack{CIFAR-10}}
            &random trigger&94.30\%&90.98\%&52.05\%&41.89364045&	0.98849749&0.020619566&0.021802150\\
            &$\mathcal{L}_{\alpha}$&94.08\%&90.72\%&92.71\%&	41.96418512&0.98918251&0.019275160&0.021482159\\
            &$\mathcal{L}_{\alpha}+\mathcal{L}_{\beta}$&95.38\%&91.06\%&	96.03\%&40.95445536&0.98685187&0.020913042&0.024609150\\
		\midrule
		&&CDA
            & SCDA&SASR&PSNR&SSIM&LPIPS&L$_1$ distance \\
			\cline{3-9}
		\multirow{3}{*}{\shortstack{GTSRB}}
            &random trigger&96.10\%&	90.14\%&48.80\%	&41.13656845&0.98806667&	0.016302601&0.013428831\\
            &$\mathcal{L}_{\alpha}$&95.38\%&91.98\%&93.12\%&	41.98251653&0.98931616&0.014626174&0.012818326\\
            &$\mathcal{L}_{\alpha}+\mathcal{L}_{\beta}$&95.06\%&92.01\%&	96.25\%	&42.16052373&	0.98821562&0.014863364&0.021438069\\
            \midrule
		&&CDA
            & SCDA&SASR&PSNR&SSIM&LPIPS&L$_1$ distance \\
			\cline{3-9}
		\multirow{3}{*}{\shortstack{CIFAR-100}}
            &random trigger&84.30\%&	92.30\%&42.19\%&41.67283447&0.98784595&0.015778990&	0.025291225\\
            &$\mathcal{L}_{\alpha}$&87.10\%&91.97\%&89.12\%&	41.56281590&0.98695216&0.014758130&0.025835158\\
            &$\mathcal{L}_{\alpha}+\mathcal{L}_{\beta}$&87.28\%&92.03\%&	96.19\%&41.22737311&0.98680114&	0.017893705&0.025383241\\
            \midrule
		&&CDA
            & SCDA&SASR&PSNR&SSIM&LPIPS&L$_1$ distance \\
			\cline{3-9}
		\multirow{3}{*}{\shortstack{Tiny ImageNet}} &random trigger&88.60\%&88.01\%&48.89\%&	42.21551066&0.98920911&0.013896998&0.012467825\\
            &$\mathcal{L}_{\alpha}$&88.08\%&91.08\%&90.93\%&	42.17145819&0.98892514&0.014215853&0.019241512\\
            &$\mathcal{L}_{\alpha}+\mathcal{L}_{\beta}$&88.10\%&91.10\%&	96.36\%&42.22344340&0.98954761&0.016416847&0.012017686\\
            \bottomrule
	\end{tabular}
	
\end{table*}

\textbf{SCDA}. In one-to-one attacks, SCDA measures whether the backdoored model can maintain prediction accuracy of clean data samples of the source label.
\begin{equation}
SCDA(f_b, \mathcal{X_S})= \frac{1}{|\mathcal{X_S}|} \sum_{{x}\in \mathcal{X_S}} \mathbf I_{[f_b({x})=y_S]},
\end{equation}
where $x$ is the clean sample of the source label dataset $\mathcal{X_S}$, and $y_S$ is the source label.

\textbf{SSIM}. SSIM is a commonly-used Quality-of-Experience (QoE) metric \cite{chen2014qos} that quantifies the differences in luminance, contrast, and structure between the original image and the distorted image.
\begin{equation}
SSIM = A(x,x')^{\alpha} B(x,x')^{\beta} C(x,x')^{\gamma},
\end{equation}
where $A(x,x'), B(x,x')$, and $C(x,x')$ quantify the luminance similarity, contrast similarity, and structural similarity between the original image $x$ and the distorted image $x'$. $\alpha, \beta$, and $\gamma$ are parameters in the range $[0, 1]$. 

\textbf{PSNR}. PSNR is computed based on MSE (Mean Squared Error) regarding the signal energy. 
\begin{equation}
\begin{split}
    PSNR = 10\log_{10} \frac{E}{MSE},\\
    MSE = \frac{1}{N} \sum_i (x_i' - x_i)^2,
\end{split}
\end{equation}
where $E$ is the maximum signal energy.

\textbf{LPIPS}. LPIPS \cite{zhang2018unreasonable} measures the similarity between two images based on the idea that the human visual system processes images in a hierarchical manner, where lower-level features, e.g., edges and textures, are processed before higher-level features, e.g., objects and scenes. The LPIPS metric uses a deep neural network to calculate the similarity between the two images. 
\begin{equation}
    LPIPS(A,B) = \sum_i w_i * ||F_i(A) - F_i(B)||^2,
\end{equation}
where $F_i(A)$ and $F_i(B)$ are the feature representations of images $A$ and $B$ at layer $i$ of the pre-trained neural network, $||.||$ denotes the L$_2$ norm, and $w_i$ is a weight that controls the relative importance of each layer. LPIPS has been shown to outperform other metrics, e.g., SSIM and PSNR, in measuring perceptual similarity between images, especially in cases where the images differ in high-level perceptual qualities such as texture and style. The smaller the value, the more similar the two images are.

\textbf{L$_1$ distance}. L$_1$ distance uses L$_1$ norm to express the distance, i.e., Manhattan distance.
\begin{equation}
    L_1~distance = \frac{1}{m}\sum_{i=1}^{m}|y_i - \check{y_i}|,
\end{equation}
where $y_i$ and $\check{y_i}$ represent the pixel values at the same position in the original image and the backdoored image, respectively, and $m$ is the size of these two images.

The larger the SSIM and the PSNR, the more similar the original and poisoned images are, and the more stealthy the trigger is. The smaller the LPIPS and the L$_1$ distance, the more stealthy the trigger is.

All experiments are implemented in Python and run on a 14-core Intel(R) Xeon(R) Gold 5117 CPU @2.00GHz  and NVIDIA GeForce RTX 2080 Ti GPU  machine running Ubuntu 18.04 system.

\begin{table*}[tt]
	\caption{Impact of trigger masking.}
	\label{tab:cut}
	\centering
        \footnotesize
        \setlength\tabcolsep{7pt}
	\begin{tabular}{lc|cccccccc}
		\toprule
&\multirow{1}{*}{\shortstack{Operation}}
           &CDA &SCDA&SASR&PSNR&SSIM&LPIPS&L$_1$ distance \\
		\cline{2-9}
		\multirow{2}{*}{\shortstack{CIFAR-10}}&No mask&96.45\%&93.99\%&98.97\%&28.03576951&0.97290265&	0.07403142&0.14443561\\
            &Mask&95.38\%&91.06\%	&96.03\%&40.95445537&0.98685187&0.02091304&0.02460915\\
		\midrule
  &\multirow{1}{*}{\shortstack{Operation}}
           &CDA &SCDA&SASR&PSNR&SSIM&LPIPS&L$_1$ distance \\
		\cline{2-9}
		\multirow{2}{*}{\shortstack{GTSRB}}&No mask&94.41\%&93.41\%&99.18\%&28.83904141&0.97496551&	0.07173179&0.09563229\\
            &Mask&95.06\%&92.01\%&96.25\%&42.16052374&0.98821563&	0.01486336&0.02143807\\
		\midrule
		&\multirow{1}{*}{\shortstack{Operation}}
           &CDA &SCDA&SASR&PSNR&SSIM&LPIPS&L$_1$ distance \\
		\cline{2-9}
		\multirow{2}{*}{\shortstack{CIFAR-100}}&No mask&87.32\%&93.77\%&99.14\%&27.01839250&0.97182516&	0.06915623&0.16583152\\
            &Mask&87.28\%&92.03\%&	96.19\%&41.22737319&0.98680114&0.01789371&0.02538324\\
            \midrule
            &\multirow{1}{*}{\shortstack{Operation}}
           &CDA &SCDA&SASR&PSNR&SSIM&LPIPS&L$_1$ distance \\
		\cline{2-9}
		\multirow{2}{*}{\shortstack{Tiny ImageNet}}&No mask&89.21\%&94.45\%&98.97\%&	27.87546510&0.97424191&	0.06857950&0.14417183\\
            &Mask&88.10\%&91.10\%&	96.36\%&42.22344340&0.98954761&0.01641684&0.01201768\\
            \bottomrule
	\end{tabular}
	\vspace{0.1cm}
\end{table*}

\begin{table*}[tt]
\vspace{0.2cm}
	\caption{Impact of model structure of ViT \cite{dosovitskiy2020image}, DeiT \cite{touvron2021training} and CaiT \cite{touvron2021going} on \sys.} 
	\label{tab:structure}
	\centering
	\setlength\tabcolsep{7pt}
        \footnotesize
	\begin{tabular}{llccccccccc}
		\toprule
		&\multirow{1}{*}{\shortstack{Structure}}
            & CDA&SCDA&SASR&PSNR&SSIM&LPIPS&L$_1$ distance \\
			\cline{2-9}
   
		\multirow{3}{*}{\shortstack{CIFAR-10}}&ViT-small&93.12\%&88.08\%&81.03\%&40.99211132&0.98661631&0.023200598&0.025389850\\
          &DeiT&95.38\%&91.06\%&96.03\%&	40.95445536&0.98685187&0.020913030&0.024609130\\
            &CaiT&90.50\%&88.83\%&88.01\%&40.37976021&0.98681241&0.028045066&0.024560293\\
		\midrule
  &\multirow{1}{*}{\shortstack{Structure}}
            & CDA&SCDA&SASR&PSNR&SSIM&LPIPS&L$_1$ distance \\
			\cline{2-9}
		\multirow{3}{*}{\shortstack{GTSRB}}&ViT-small&92.41\%&82.32\%&89.30\%&42.47531969&0.99082177&0.018190213&0.018013531\\
            &DeiT&95.06\%&92.01\%&96.25\%&	42.16052370&0.98821562&0.014863364&0.021438069\\
            &CaiT&93.09\%&82.31\%&80.14\%&41.67578074&0.99008435&0.018726882&0.020281965\\
            \midrule
  &\multirow{1}{*}{\shortstack{Structure}}
            & CDA&SCDA&SASR&PSNR&SSIM&LPIPS&L$_1$ distance \\
			\cline{2-9}
		\multirow{3}{*}{\shortstack{CIFAR-100}}&ViT-small&84.64\%&89.09\%&87.18\%&41.69399703&0.98854577&0.017296846&0.023896610\\
        &DeiT&87.28\%&92.03\%&96.19\%&41.22737319&	0.98680115&
            0.017893705&0.025383241\\
            &CaiT&84.95\%&91.97\%&78.99\%&40.98745143&0.98799694&0.021582731&0.026102297\\
            \midrule
            &\multirow{1}{*}{\shortstack{Structure}}
            & CDA&SCDA&SASR&PSNR&SSIM&LPIPS&L$_1$ distance \\
			\cline{2-9}
		\multirow{3}{*}{\shortstack{Tiny ImageNet}}&ViT-small&87.25\%&87.98\%&92.31\%&41.30535876&0.98899472&
            0.020672306&0.024818577\\
            &DeiT&88.10\%&91.10\%&96.36\%&42.22344340&0.98954761&0.016416847&0.012017686\\
            &CaiT&87.67\%&81.90\%&91.34\%&43.41107503&0.99168205&0.015393089&0.017397709\\
            \bottomrule
	\end{tabular}
	
\end{table*}
\section{Evaluation Results}

\subsection{Comparison with State-of-the-art Baselines}
We compare \sys with five state-of-the-art vision transformer backdoor attacks, i.e., DBIA \cite{lv2021dbia}, BAVT \cite{subramanya2022backdoor},  DBAVT-BadNets \cite{doan2022defending}, TrojViT \cite{zheng2022trojvit}, and DBAVT-WaNet \cite{doan2022defending}. As the source code of BadViT \cite{yuan2023you} is not publicly available, we did not compare \sys with it. 
We implement the baseline attacks using their published source codes. 

\textbf{One-to-one attacks.} 
In the one-to-one attack, we choose a specific label as the source label and then construct poisoned samples in that label. The comparison results are shown in Table~\ref{tab:com}.
Compared with visible trigger backdoor attacks (i.e., DBIA, DBAVT-BadNets, and TrojViT), \sys demonstrates a comparable attack success rate but outperforms baseline methods across all four datasets in terms of image quality metrics. Notably, \sys achieves significantly higher PSNR, lower LPIPS, and lower L$_1$ distance. For example, \sys achieves PSNR of 40.9544, 42.1605, 41.2273, and 42.2234 on CIFAR-10, GTSRB, CIFAR-100, and Tiny ImageNet models, respectively, while the highest PSNR of the baseline is 24.0267 (CIFAR-10), 23.5182 (GTSRB), 22.9964 (CIFAR-100), and 23.9338 (Tiny ImageNet).

Compared with hidden trigger backdoor attacks (i.e., BAVT and DBAVT-WANET), \sys achieves the highest success rate across all datasets. The ASR values for \sys are 96.03\%, 96.25\%, 96.19\%, and 96.36\% for CIFAR-10, GTSRB, CIFAR-100, and Tiny ImageNet, respectively. In contrast, BAVT achieves lower attack success rates of 80.33\%, 82.15\%, 78.38\%, and 87.22\%, followed by DBAVT-WaNet with ASR values of only 19.67\%, 4.64\%, 9.82\%, and 15.29\%. Notably, BAVT's inability to achieve high ASR is primarily attributed to its lack of a trigger optimization process. The ineffectiveness of DBAVT-WaNet in vision transformers is also substantiated in \cite{subramanya2023vision}.

We also present the poisoned samples of baselines and \sys in Fig.~\ref{fig:com}. We can see that the poisoned samples generated by \sys are quite similar to the benign samples and are easy to evade human visual inspections. The poisoned samples of BAVT are natural in most cases, but there are still noises visible to the human eye in some cases. 
Although DBAVT-WaNet can generate relatively natural backdoor samples, it cannot achieve a high attack success rate. 

\begin{table*}[tt]
	\caption{Impact of trigger location on \sys. } 
	\label{tab:location}
	\centering
        \footnotesize
        \setlength\tabcolsep{7pt}
	\begin{tabular}{lc|ccccccccc}
		\toprule&\multirow{1}{*}{\shortstack{Position}}
          & CDA & SCDA&SASR&PSNR&SSIM&LPIPS&L$_1$ distance \\
			\cline{2-9}
		
		\multirow{5}{*}{\shortstack{CIFAR-10}}&(20,20)&96.60\%&89.91\%&97.01\%
            &40.86427487&0.98663920&0.034751944&0.025843668\\
            &(40,40)&95.07\%&89.00\%&97.00\%
            &40.66736227&0.98725157&0.045461105&0.025837701\\
            &(60,60)&95.90\%&88.00\%&96.99\%
            &40.84064891&0.98880815&0.043541587&0.025003017\\
            &(80,80)&95.67\%&88.91\%&97.00\%
            &40.61767301&0.98845237&0.043229166&0.027704255\\
            &(100,100)&96.22\%&89.72\%&97.12\%
            &40.73122526&0.98947763&0.040025223&0.026596570\\
		\midrule
          &\multirow{1}{*}{\shortstack{Position}}
          & CDA & SCDA&SASR&PSNR&SSIM&LPIPS&L$_1$ distance \\
			\cline{2-9}
		\multirow{5}{*}{\shortstack{GTSRB}}&(20,20)&94.04\%&91.37\%&95.57\%&40.39302398&0.99186468&	0.022288310&0.029558085\\
            &(40,40)&93.04\%&91.00\%&96.00\%&41.01812721&	0.99302870&0.024230021&0.028118778\\
            &(60,60)&92.11\%&91.00\%&95.99\%&42.63558145&	0.99434155&0.014624223&0.019823201\\
            &(80,80)&92.09\%&91.73\%&96.00\%&41.41952619&	0.99540782&0.029697580&0.025249460\\
            &(100,100)&94.22\%&92.47\%&97.10\%&40.77952090&	0.99259483&0.029544420&0.026000121\\
		\midrule
          &\multirow{1}{*}{\shortstack{Position}}
          & CDA & SCDA&SASR&PSNR&SSIM&LPIPS&L$_1$ distance \\
			\cline{2-9}
		\multirow{5}{*}{\shortstack{CIFAR-100}}&(20,20)&87.10\%&92.97\%&95.76\%&41.47786213&0.98515772&	0.016551531&0.025660211\\
            &(40,40)&87.95\%&94.00\%&95.00\%&41.61916425&	0.98401403&0.017193300&0.020908031\\
            &(60,60)&87.63\%&93.00\%&94.23\%&41.24616020&	0.98726010&0.016123900&0.021682334\\
            &(80,80)&87.16\%&94.53\%&98.00\%&41.45717757&	0.98232448&0.016496101&0.021554891\\
            &(100,100)&86.48\%&94.24\%&96.88\%&41.68171064&	0.98221159&0.019261500&0.022997227\\
            \midrule
           &\multirow{1}{*}{\shortstack{Position}}
          & CDA & SCDA&SASR&PSNR&SSIM&LPIPS&L$_1$ distance \\
			\cline{2-9}
		\multirow{5}{*}{\shortstack{Tiny ImageNet}}&(20,20)&87.27\%&91.61\%&96.05\%
            &41.53427867&0.98901247&0.021507127&0.021378787\\
            &(40,40)&87.69\%&91.00\%&96.05\%
            &41.86912918&0.99049633&0.024181865&0.021287050\\
            &(60,60)&88.34\%&91.28\%&94.90\%
            &42.02730122&0.99139136&0.017556518&0.011976972\\
            &(80,80)&86.81\%&91.98\%&96.27\%
            &41.24157382&0.99078094&0.022223294&0.019934125\\
            &(100,100)&88.38\%&91.56\%&96.56\%
            &41.68395828&0.99072712&0.024962762&0.017789589\\
            \bottomrule
	\end{tabular}
	
\end{table*}

\begin{table}[tt]
\vspace{0.1cm}
	\caption{Impact of trigger location change.}

	\label{tab:attentiondiff2}
	\centering
	\footnotesize
	\setlength\tabcolsep{6pt}
	\begin{tabular}{c|cccccccc}
		\toprule
		Token number& CIFAR-10&GTSRB& CIFAR-100& Tiny ImageNet \\
		\midrule
            0&96.17\%&95.05\%&96.19\%&96.36\%\\
            2&95.34\%&96.25\%&93.41\%&92.12\%\\
            4&90.00\%&90.17\%&85.98\%&93.34\%\\
            6&87.21\%&86.64\%&85.24\%&86.01\%\\
            8&84.53\%&85.98\%&85.01\%&85.20\%\\
		\bottomrule
	\end{tabular}
 \vspace{-0.3cm}
\end{table}

\begin{table*}[tt]
	\caption{Impact of poison rate on \sys. } 
	\label{tab:ratio}
	\centering
	\setlength\tabcolsep{7pt}
	\footnotesize
	\begin{tabular}{lc|ccccccccc}
		\toprule
          &\multirow{1}{*}{\shortstack{Rate}}
          & CDA & SCDA&SASR&PSNR&SSIM&LPIPS&L$_1$ distance \\
		\cline{2-9}
		\multirow{5}{*}{\shortstack{CIFAR-10}}&2\%&93.20\%&94.97\%&77.93\%&	40.57049098&0.98624932&0.024620646&0.025942143\\
            &4\%&92.48\%&94.06\%	&84.09\%&40.95445537&0.98685187&0.020913040&0.024609150\\
            &6\%&92.84\%&94.97\%&89.97\%&
        40.84064891&0.98685187&0.043541587&0.025003017\\
            &8\%&93.04\%&93.20\%&93.03\%&
        40.67720354&0.98607736&0.025473694&0.026723464\\
            &10\%&92.89\%&93.32\%&94.46\%&
        41.11273874&0.98685663&0.021309917&0.024547774\\
		\midrule
   &\multirow{1}{*}{\shortstack{Rate}}
          & CDA & SCDA&SASR&PSNR&SSIM&LPIPS&L$_1$ distance \\
		\cline{2-9}
		\multirow{5}{*}{\shortstack{GTSRB}}&2\%&92.77\%&	94.78\%&77.96\%&40.29474451&0.99200534&0.022160474&	0.025118071\\
            &4\%&93.09\%&92.01\%&84.89\%&40.44011445&	0.99058669&0.023885502&0.027037770\\
            &6\%&93.68\%&93.88\%&87.00\%&42.16052374&0.98821563&	0.014863360&0.021438070\\
            &8\%&92.27\%&92.96\%&94.02\%&40.94053573&	0.98978978&0.022004828&0.028205888\\
            &10\%&93.27\%&93.00\%&95.98\%&39.49662207&	0.99045169&0.027001012&0.031210689\\
		\midrule
   &\multirow{1}{*}{\shortstack{Rate}}
          & CDA & SCDA&SASR&PSNR&SSIM&LPIPS&L$_1$ distance \\
		\cline{2-9}
		\multirow{5}{*}{\shortstack{CIFAR-100}}&2\%&86.96\%&89.92\%&88.90\%&41.82908167&	0.98505330&0.017630098&0.023788085\\
            &4\% &86.46\%&91.96\%&93.77\%&41.85021525&
        0.98449587&0.015876947&0.022336865\\
            &6\% &86.28\%&92.03\%&96.18\%&41.22737319&
        0.98460483&0.017893713&0.025383241\\
            &8\% &86.46\%&92.54\%&96.92\%&41.90455727&
        0.98781959&0.015303540&0.021567119\\
            &10\% &86.63\%&93.00\%&97.01\%&41.60080208&
        0.98480918&0.015680369&0.025409516\\
           \midrule
            &\multirow{1}{*}{\shortstack{Rate}}
          & CDA & SCDA&SASR&PSNR&SSIM&LPIPS&L$_1$ distance \\
		\cline{2-9}
		\multirow{5}{*}{\shortstack{Tiny ImageNet}}&2\%&87.29\%&92.20\%&79.98\%
        &42.44245681&0.99093413&0.013690100&0.014958851\\
            &4\%&88.01\%&93.03\%&89.67\%
        &42.35118580&0.99095189&0.014618013&0.018686902\\
            &6\%&86.80\%&92.65\%&	95.36\%&42.22344340&0.98954761&0.016416840&0.012017686\\
            &8\%&87.32\%&90.34\%&95.44\%
        &42.31759820&0.99095189&0.014258561&0.011420765\\
            &10\%&86.40\%&90.68\%&96.57\%
        &42.38477341&0.98986750&0.013991630&0.018679644\\
            \bottomrule
	\end{tabular}
	\vspace{0.1cm}
\end{table*}

\textbf{Any-to-one attacks. }
In the any-to-one attack, we aim to mislead the backdoored model by misclassifying any sample containing the trigger into the specified target class. 
To conduct the any-to-one attacks, we choose any label except the target label as the source label and proceed with both trigger generation and the poisoned sample generation processes in \sys. As shown in Table~\ref{tab:com2}, \sys consistently achieves a high attack success rate ($\geq 95\%$) and maintains good image quality across all four datasets under this setting.

\subsection{Ablation Study}

\textbf{Impact of trigger generation loss}.
We examine the necessity of the two loss items, i.e., $\mathcal{L}_{\alpha}$ and $\mathcal{L}_{\beta}$, in the trigger generation process. The results are shown in Table~\ref{tab:ablation-trigger generation}. Compared with random triggers, we can observe that $\mathcal{L}_{\alpha}$ can significantly improve ASR. For example, for GTSRB dataset, the ASR of using random trigger is 48.80\%, but reaches as high as 93.12\% using the $\mathcal{L}_{\alpha}$ loss function. The increment is more than 44.32\%. Similarly, for Tiny ImageNet, the improvement in ASR is more than 42.04\% with $\mathcal{L}_{\alpha}$. The success of $\mathcal{L}_{\alpha}$ lies in its ability to minimize the distance between the poisoned sample and the sample of the target label with respect to each layer's attention, thereby aiding the model in misclassifying the backdoored samples to the target label. Compared with the $\mathcal{L}_{\alpha}$ loss function, the $\mathcal{L}_{\alpha}+\mathcal{L}_{\beta}$ attack further increases ASR, where the average improvement of the four datasets exceeded 4.74\%. The success of $\mathcal{L}_{\beta}$ lies in its ability to increase the importance of the attention diffusion area during training.

\textbf{Impact of trigger masking}.
We investigate the impact of the trigger masking process on the attack performance of \sys.
The results are presented in Table~\ref{tab:cut}.
It is demonstrated that the \sys with mask method consistently outperforms \sys with no-mask regarding image quality metrics across all datasets. Furthermore, the attack success rate of \sys with the mask is only marginally lower than that of \sys without the mask.


\textbf{Impact of model structure.} 
By default, we assume the victim user employs the DeiT model structure \cite{touvron2021training} to train ViT models on CIFAR-10, GTSRB, CIFAR-100, and Tiny ImageNette. In this section, we investigate the efficacy of \sys when the victim user adopts other model architectures, including ViT-small \cite{touvron2021training} and CaiT \cite{touvron2021going}.
The results are shown in Table~\ref{tab:structure}. The results demonstrate the robustness of \sys to various user-end vision transformer architectures. Notably, \sys can successfully inject backdoors into the target vision transformer regardless of its specific structure.

\begin{table*}[tt]
	\caption{{Impact of the number of sub-trigger on \sys.}} 
	\label{tab:size2}
	\centering
        \footnotesize
        \setlength\tabcolsep{7pt}
	\begin{tabular}{lc|ccccccccc}
 	\toprule
		 &\multirow{1}{*}{\shortstack{Number}}
          & CDA & SCDA&SASR&PSNR&SSIM&LPIPS&L$_1$ distance \\
		\cline{2-9}
		\multirow{5}{*}{\shortstack{CIFAR-10}}
            &2&97.34\%&92.14\%&97.01\%&34.78335644&0.98133081&	0.049309037&	0.049309037\\
            &4&97.21\%&91.31\%&96.08\%&37.92072323&0.98486185&0.030391748&	0.035931080\\
            &8&96.38\%&91.06\%&96.03\%&40.95445537&0.98685187&0.020913042&	0.024609150\\
            &16&96.12\%&91.90\%&95.10\%&43.36271992&0.98752760&0.018397378&	0.021207634\\
            &32&95.95\%&92.89\%&93.93\%&45.27171766&0.98915138&0.016679952&	0.019112436\\
		\midrule
   &\multirow{1}{*}{\shortstack{Number}}
          & CDA & SCDA&SASR&PSNR&SSIM&LPIPS&L$_1$ distance \\
		\cline{2-9}
		\multirow{5}{*}{\shortstack{GTSRB}}
            &2&96.04\%&93.89\%&98.31\%&36.12928465&
            0.98192456&0.028236278&0.044755183\\
            &4&95.27\%&93.45\%&97.97\%&39.18887023&	0.98614758&0.021927478&0.028420347\\
            &8&95.06\%&92.01\%&96.25\%&42.16052374&0.98821563&	0.014863360&0.021438070\\
            &16&93.85\%&91.97\%&89.72\%&43.45372817&
            0.99096673&0.018397378&0.020738195\\
            &32&92.05\%&93.02\%&85.46\%&45.27171766&
            0.99221992&0.016738295&0.019132648\\
		\midrule
   &\multirow{1}{*}{\shortstack{Number}}
          & CDA & SCDA&SASR&PSNR&SSIM&LPIPS&L$_1$ distance \\
		\cline{2-9} 
		\multirow{5}{*}{\shortstack{CIFAR-100}}
            &2&88.91\%&94.53\%&98.96\%&35.28076779&	0.99998784&0.032241136&0.054148361\\
            &4&87.65\%&92.95\%&96.74\%&38.57694349&
            0.99999475&0.027638825&0.033707149\\
            &8&87.28\%&92.03\%&96.19\%&41.22737319&
            0.99999694&0.017893705&0.025383241\\
            &16&86.34\%&92.97\%&95.44\%&43.48291515&
            0.99166753&0.018474821&0.021206731\\
            &32&85.27\%&93.18\%&92.01\%&45.27185215&
            0.99389779&0.016677821&0.019112521\\
            \midrule
             &\multirow{1}{*}{\shortstack{Number}}
          & CDA & SCDA&SASR&PSNR&SSIM&LPIPS&L$_1$ distance \\
		\cline{2-9}
		\multirow{5}{*}{\shortstack{Tiny ImageNet}}
            &2&88.40\%&96.02\%&97.98\%&35.11170010&0.98237890&0.036974303&0.030241988\\
            &4&86.80\%&94.51\%&96.70\%&39.16717749&0.98843812&0.018885103&0.021401446\\
            &8&88.10\%&91.10\%&	96.36\%&42.22344340&0.98954761&0.016416840&0.012017686\\
            &16&87.20\%&95.78\%&89.75\%&44.85323547&0.99117594&0.012424159&0.009624915\\
            &32&87.20\%&92.24\%&82.18\%&45.33753916&0.99143141&0.012572141&0.009514139\\
            \bottomrule
	\end{tabular}

\end{table*}

\begin{table*}[tt]
	\caption{Impact of $\gamma$ on \sys. } 
	\label{tab:gamma}
	\centering
	\setlength\tabcolsep{7pt}
 \footnotesize
	\begin{tabular}{lc|ccccccccc}
		\toprule
 &\multirow{1}{*}{\shortstack{Value}}
          & CDA & SCDA&SASR&PSNR&SSIM&LPIPS&L$_1$ distance \\
		\cline{2-9}
		\multirow{5}{*}{\shortstack{CIFAR-10}}&0&95.12\%&85.72\%&81.98\%&	41.14551097&0.98685187&0.020549812&0.024040443\\
            &0.1&95.05\%&87.06\%&92.04\%&40.89031788&0.98696815&0.021580349&	0.024613553\\
            &1&95.38\%&91.06\%&96.03\%&40.95445537&0.98685187&0.020913040&0.024609150\\
            
            &10&95.10\% &88.34\%&96.27\%&40.96945439&0.98686838&0.022295979&	0.024724841\\
            &100&95.47\%&87.99\%&94.34\%&40.94612039&0.98688238&0.022119063&	0.024915346\\
		\midrule
   &\multirow{1}{*}{\shortstack{Value}}
          & CDA & SCDA&SASR&PSNR&SSIM&LPIPS&L$_1$ distance \\
		\cline{2-9}
		\multirow{5}{*}{\shortstack{GTSRB}}&0&92.31\%&	88.02\%&87.00\%&40.61745879&0.99075325&0.019963886&	0.027210642\\
            &0.1&93.27\%&90.98\%&93.98\%&41.35280142&	0.99084526&0.018742824&0.026673838\\
            &1  &95.06\%&92.01\%&96.25\%&42.16052374&0.98821563&	0.014863360&0.021438070\\
            &10 &95.13\%&91.12\%&87.03\%&43.09401597&	0.99103021&0.015680834&0.018582381\\
            &100&93.36\%&89.93\%&84.49\%&41.05788029&	0.98982465&0.017147312&0.025486335\\
		\midrule
   &\multirow{1}{*}{\shortstack{Value}}
          & CDA & SCDA&SASR&PSNR&SSIM&LPIPS&L$_1$ distance \\
		\cline{2-9}
		\multirow{5}{*}{\shortstack{CIFAR-100}}&0&87.42\%&88.89\%&90.03\%&41.72688963&0.98460483&
               0.016448218&0.023634949\\
           &0.1&87.61\%&90.12\%&91.31\%&41.77645908&0.98514050&	0.017881361&0.025709828\\
            &1&87.28\%&92.03\%&96.19\%&41.22737319&0.98680114&0.017893710&0.025383240\\
            &10 &87.30\%&92.67\%&93.34\%&41.81291517&0.98430532&
               0.016330981&0.025761703\\
            &100&87.46\%&90.45\%&85.56\%&41.95378033&0.98490804& 
               0.016251433&0.025466429\\
           \midrule
            &\multirow{1}{*}{\shortstack{Value}}
          & CDA & SCDA&SASR&PSNR&SSIM&LPIPS&L$_1$ distance \\
		\cline{2-9}
		\multirow{5}{*}{\shortstack{Tiny ImageNet}}
              &0&88.38\%&83.76\%&88.00\%&
            42.35118581&0.98954761&0.014125096&0.018686903\\
            &0.1&87.99\%&87.53\%&91.41\%&	42.37398265&0.98991251&0.015848395&0.017364718\\
            &1&88.10\%&91.10\%&	96.36\%&42.22344340&0.98954761&0.016416840&0.012017686\\
            &10&86.81\%&88.13\%&93.74\%&	42.35118581&0.99051257&0.014125096&0.015050205\\
            &100&86.45\%&87.28\%&92.48\%&	42.35118581&0.99082915&0.013690100&0.018686905\\
            \bottomrule
	\end{tabular}
	
\end{table*}

\begin{table*}[tt]
	\caption{Impact of $\phi_A$ transparency value on \sys.} 
	\label{tab:trans}
	\centering
	\footnotesize
     \setlength\tabcolsep{7pt}
	\begin{tabular}{lc|ccccccccc}
		\toprule
  
        &\multirow{1}{*}{\shortstack{Transparency}}
	&CDA & SCDA&SASR&PSNR&SSIM&LPIPS&L$_1$ distance \\
		\cline{2-9}
	\multirow{6}{*}{{CIFAR-10}}&
		  1&  94.10\%&94.20\%&98.69\%&37.66517583&0.98474550&0.03388371&0.03086529\\
            &0.7&96.25\%	&94.24\%&97.75\%&38.93397673&0.98551257&0.02949779&0.02737223\\
            &0.6&95.99\%&92.10\%&97.24\%&39.95977355&0.98572815&0.02284010&	0.02637787\\
            &0.5&95.40\%&91.17\%&96.17\%&40.88510985&0.98685187& 0.02346396&0.02768558\\
            &0.4&95.04\%&92.23\%&94.86\%&41.72560621&0.98875823&  0.02278105&	0.02472355\\
            &0.3&94.91\%&93.81\%&93.80\%&43.24908694&0.98893156&  0.01612502&	0.02252965\\
		\midrule
		
         &\multirow{1}{*}{\shortstack{Transparency}}
	&CDA & SCDA&SASR&PSNR&SSIM&LPIPS&L$_1$ distance \\
		\cline{2-9}
  \multirow{6}{*}{{GTSRB}}&
          1&94.69\%&91.98\%&98.89\%&38.03517442&0.98772788&0.03123480&0.02941255\\
           & 0.7&95.24\%&92.00\%&96.81\%&42.00091317&0.99028314&0.01715673&0.01753078\\
            &0.6&95.10\%&92.00\%&96.56\%&41.91790828&0.99063527&0.01598372&0.02036995\\
            &0.5&95.06\%&92.01\%&96.25\%&42.16052373&0.99093508&0.01486336&0.02143806\\
            &0.4&94.85\%&92.03\%&93.47\%&44.19512798&0.99115485&0.01269197&0.01842885\\
            &0.3&94.01\%&91.87\%&89.46\%&45.72819657&0.99143615&0.01182940&0.01573821\\
           \midrule
		
		&\multirow{1}{*}{\shortstack{Transparency}}
	&CDA & SCDA&SASR&PSNR&SSIM&LPIPS&L$_1$ distance \\
		\cline{2-9}
		\multirow{6}{*}{{CIFAR-10}}&
            1&87.65\%&93.00\%&99.12\%&37.14010549&0.98554849&0.02812124&0.03364957\\
            &0.7&87.50\%&93.00\%&97.12\%&40.28493102&	0.98403257&0.02115619&0.03121815\\
            &0.6&87.38\%&93.00\%&96.68\%&41.47741762&	0.98508405& 0.01978048&0.02855978\\
            &0.5&87.98\%&92.46\%&96.21\%&41.91068052&	0.98680114&0.01744731&0.02556451\\
            &0.4&87.27\%&92.13\%&94.25\%&42.53917959&	0.98719325&0.01574666&0.02230548\\
            &0.3&87.14\%&92.77\%&93.21\%&43.18378141&	0.98799734&0.01241575&0.02141782\\
           \midrule
		
		
  	&\multirow{1}{*}{\shortstack{Transparency}}
	&CDA & SCDA&SASR&PSNR&SSIM&LPIPS&L$_1$ distance \\
		\cline{2-9}
  \multirow{6}{*}{{Tiny ImageNet}}&
            1&87.65\%&92.97\%&98.37\%&38.30271790&0.98465052&0.02769978&0.02733095\\
            &0.7&84.80\%&93.14\%&	96.90\%&40.25917212&	0.98894382&	0.01776854&	0.01793515\\
            &0.6&84.80\%&92.67\%&	96.50\%&41.07829954&0.98921541&	0.01614780&	0.01284087\\
            &0.5&88.40\%&91.10\%&96.36\%&42.03038703&	0.98954761&0.01653650&0.01244495\\
            &0.4&86.00\%&92.66\%&93.79\%&43.24320262&	0.99066859&	0.01416404&	0.01212599\\
            &0.3&87.60\%&92.82\%&88.30\%&44.74291306&	0.99184259&	0.01059088&	0.01023500\\
            \bottomrule
	\end{tabular}
	\vspace{0.1cm}
\end{table*}
\begin{figure*}[tt]
	\centering
	\begin{minipage}[t]{3.22in}
		\centering
		\includegraphics[trim=0mm 0mm 0mm 0mm, clip,width=3.22in]{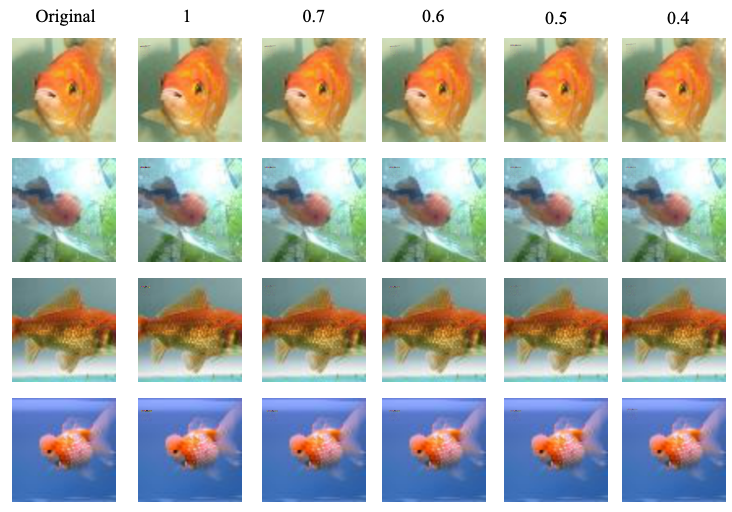}\\
		\centerline{\small (a) $\phi_A$}
	\end{minipage}
	\hspace{-0cm}
	\begin{minipage}[t]{3.25in}
		\centering
		\includegraphics[trim=0mm 0mm 0mm 0mm, clip,width=3.25in]{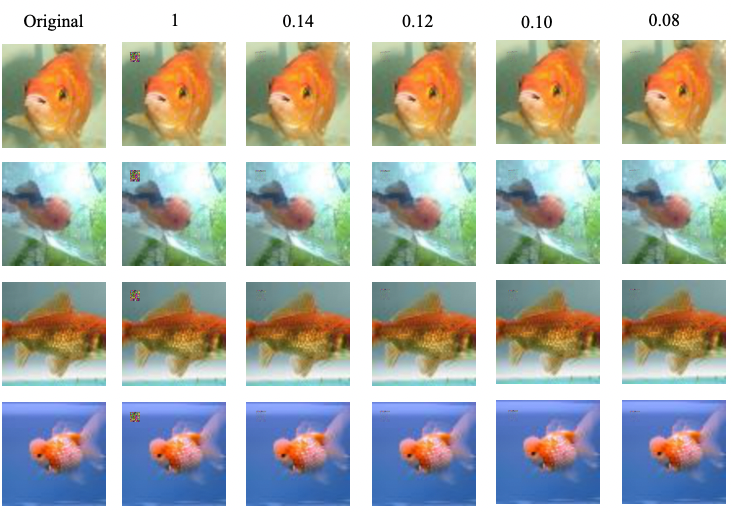}\\
		\centerline{\small (b) $\phi_D$}
	\end{minipage}
	\caption{Impact of transparency value of $\phi_A$ and $\phi_D$.} \label{fig:trans}
\end{figure*}

\begin{table*}[tt]
	\caption{Impact of $\phi_D$ transparency value on \sys.} 
	\label{tab:trans2}
	\centering
	\setlength\tabcolsep{5pt}
 \footnotesize
	\begin{tabular}{lc|cccccccc}
		\toprule
		&\multirow{1}{*}{\shortstack{Transparency}}
	&CDA & SCDA&SASR&PSNR&SSIM&LPIPS&L$_1$ distance \\
		\cline{2-9}
	\multirow{6}{*}{{CIFAR-10}}&
		1&95.90\%&91.19\%&99.01\%&26.43469929&0.97166514&0.084028404&0.17961293\\
            &0.14&95.77\%&93.03\%&	97.24\%&	39.66574432&	0.98422473&	 
             0.022998237&	0.03013849\\
            &0.12&95.49\%&91.12\%	&97.02\%&   40.25672329&    0.98538231&	0.023463968&	0.02768558\\
            &0.10&95.40\%&90.98\%	&96.17\%&	40.88510985&   0.98685187	&0.022781052&	0.02472356\\
            &0.08&95.18\%&90.21\%&95.99\%	&   41.29501765&	0.98819245
            &0.018896166&	0.02162306 \\
            &0.06&94.95\%&91.96\%	&93.87\%&	  42.02247015&0.98852146&	 
             0.020259786&	0.01831400\\
		\midrule
  	&\multirow{1}{*}{\shortstack{Transparency}}
	&CDA & SCDA&SASR&PSNR&SSIM&LPIPS&L$_1$ distance \\
		\cline{2-9}
		 \multirow{6}{*}{{GTSRB}}&
             1&95.03\%&93.10\%&99.97\%&27.87791642&0.97418659&0.064235692&0.14239646\\
            &0.14&95.41\%&93.38\%&97.24\%&40.67418291&
            0.98842328&0.018429141&0.02738194\\
            &0.12&95.30\%&92.14\%&96.98\%&41.62877339&	0.98995128&0.016425582&0.02349806\\
            &0.10&95.06\%&92.01\%&96.25\%&42.16052373&
            0.99093508&0.014863364&0.02143806\\
            &0.08&94.99\%&92.04\%&92.50\%&43.40743151&
            0.99140088&0.011079715&0.01618935\\
            &0.06&94.87\%&93.05\%&82.09\%&45.99887800&
            0.99162751&0.006269692&0.01033486\\

           \midrule
		
        &\multirow{1}{*}{\shortstack{Transparency}}
	&CDA & SCDA&SASR&PSNR&SSIM&LPIPS&L$_1$ distance \\
		\cline{2-9}
		 \multirow{6}{*}{{CIFAR-100}}&
            1&87.58\%&93.12\%&99.85\%&27.55692235&0.97323906&0.072205021&0.14881321\\
            &0.14&87.45\%&93.37\%&96.86\%&39.76552327&	0.98472261&0.023346093&0.03429951\\
            &0.12&87.43\%&92.14\%&96.41\%&40.36324470&	0.98587515&0.019122116&0.02872732\\
            &0.10&87.28\%&92.03\%&96.19\%&41.22737319&	0.98680114&0.017893705&0.02538324\\
            &0.08&87.72\%&91.91\%&95.11\%&41.70913533&	0.98925381&0.016675860&0.02287852\\
            &0.06&87.49\%&92.89\%&93.14\%&41.90908652&	0.99098426&0.016560421&0.01960295\\
            \midrule
           
        &\multirow{1}{*}{\shortstack{Transparency}}
	&CDA & SCDA&SASR&PSNR&SSIM&LPIPS&L$_1$ distance \\
		\cline{2-9}
		\multirow{6}{*}{{Tiny ImageNet}}&
           1&88.41\%&92.94\%&99.41\%&27.44181815&0.97289526&0.068457670&0.15543706\\
            &0.14&89.20\%&93.88\%&97.99\%&40.95360854&	0.98758392&	0.019566971&0.01557544\\
            &0.12&87.60\%&92.31\%&97.10\%&41.44975893&	0.98899579&	0.016794591&0.02281016\\
            &0.10&88.10\%&91.10\%&96.36\%&42.22344342&  0.98954761&	0.016416847&0.01201768\\
            &0.08&88.80\%&92.58\%&93.69\%&42.90084113&	0.99222844&0.012737656&0.01595382\\
            &0.06&84.40\%&93.01\%&92.10\%&43.25995909&	0.99269813&	0.013122240&0.01103902\\
            \bottomrule
	\end{tabular}
\end{table*}

\textbf{{\color{black}Impact of trigger location in the source image.}}
We examine how the attack success rate varies when the trigger is positioned at different locations within the source image. It's important to note that during the testing process, the sub-trigger's location matches that of the patched samples.
As shown in Table~\ref{tab:location}, $x$ and $y$ represent the horizontal and vertical coordinate values of the plane Cartesian coordinate system formed by extending left and downward from the upper left corner of the image as the origin. We can see that the attack performance of \sys is robust to the location of the trigger. 


\textbf{Impact of trigger location change.}
To achieve the best attack performance, the sub-trigger should be in the same position in the training and testing phase.
We now evaluate the change of sub-trigger locations in the testing procedure. During testing, we put the sub-trigger within the attention diffusion area but not exactly the location in the generated patched samples.
As shown in Table~\ref{tab:attentiondiff2}, the SASR will decrease as the diffusion area expands since the sub-trigger deviates more from the location at the victim ViT training phase. However, \sys can also maintain a high attack success rate ($\geq 85\%$) when the attention diffusion area is six tokens, which is much larger than the trigger area.

\textbf{Impact of poison rate.} The poison rate refers to the proportion of poisoned samples to all training samples. In our study, we investigate the impact of poison ratio as shown in Table~\ref{tab:ratio}. The results demonstrate that as the poison rate increases, the attack success rate also tends to rise. For instance, when the poison ratio is set at 2\%, \sys achieves an attack success rate of 77.93\% for CIFAR-10 and 77.96\% for GTSRB. However, with a higher poison ratio of 10\%, \sys achieves higher ASRs of 94.46\% for CIFAR-10 and 95.98\% for GTSRB. Notably, \sys maintains a consistently high model prediction accuracy and image quality even as the poison rate increases. 


\textbf{Impact of the number of sub-trigger.}
We explore the impact of the number of sub-triggers on the attack performance of \sys. 
Note that the number of sub-triggers corresponds to the number of areas in the trigger masking process. However, the number of triggers on the patched image is always one. The results are shown in Table~\ref{tab:size2}. 
Before the sub-trigger number is reduced to 8, as the sub-trigger number decreases, the attack success rate remains largely unchanged, but the sub-trigger becomes less visible with better image quality metrics. With a sub-trigger number of 8, \sys achieves ASR of 96.03\% (CIFAR-10), 96.25\% (GTSRB), 96.19\% (CIFAR-100), and 96.36\% (Tiny ImageNet) and achieves PNSR of 40.9544 (CIFAR-10), 42.1605 (GTSRB), 41.2273 (CIFAR-100), and 42.2234 (Tiny ImageNet). 
In \sys, we set the sub-trigger number as 8.

\textbf{Impact of $\gamma$.} We explore the impact of the hyperparameter $\gamma$ in the loss function on the attack performance. The results are shown in Table~\ref{tab:gamma}. It is observed that when the value of $\gamma$ is set to 1, the attack success rate and clean data accuracy consistently reach their optimal values in the majority of cases. In the experiments, we set $\gamma$ as 1 by default.



\textbf{Impact of transparency adjustment.}
There are two parameters in transparency adjustment, i.e., $\phi_A$ and $\phi_D$. $\phi_A$ is the transparency of the sub-trigger area, and $\phi_D$ is the transparency of the rest trigger area. The lower the transparency value is, the more imperceptible the trigger is. We explore the impact of the transparency value of $\phi_A$ in Table~\ref{tab:trans} and the impact of the transparency value of $\phi_D$ in Table~\ref{tab:trans2}. We also present the backdoored samples with different transparency values in Fig.~\ref{fig:trans}. We can see that as the transparency value is smaller, the trigger is more concealed, and the backdoored samples have better image qualities. However, the attack success rate will be decreased. When the transparency value of $\phi_A$ and $\phi_D$ are set as 0.5 and 0.1, the human eyes can hardly discern the trigger. Thus, in the experiments, we set the $\phi_A$ as 0.5 and $\phi_D$ as 0.1 by default.


\subsection{Time Cost}
To evaluate the efficiency of \sys, we conducted a comparison of computational costs between \sys and the baselines, as shown in Table~\ref{tab:time}. We can see that \sys demonstrates similar computational costs to the baseline attacks. Importantly, \sys achieves superior attack performance while maintaining reasonable computational efficiency.

\section{Robustness to State-of-the-art Backdoor Defenses}
In this section, we investigate the ability of \sys to evade state-of-the-art backdoor defenses. Currently, there are only two available defenses specifically designed for vision transformers, which are DBAVT \cite{doan2022defending} and BAVT \cite{subramanya2022backdoor}. We also consider adapting three other advanced backdoor defenses, namely Beatrix \cite{ma2022beatrix}, TeCo \cite{liu2023detecting}, and SAGE \cite{23spgong}, to defend against \sys.

\textbf{DBAVT. } DBAVT \cite{doan2022defending} mitigates the backdoor attacks on ViTs using patch processing. It is based on the insight that clean-data accuracy and backdoor attack success rates of ViTs respond differently to patch transformations before the positional encoding, unlike CNN models. 

As shown in Table~\ref{tab:dbavt}, after applying DBAVT, the ASR of \sys only slightly decreases.
\sys can also achieve ASR of 96.05\%, 81.34\%, 100\%, and 94.89\% for CIFAR-10, GTSRB, CIFAR-100, and Tiny ImageNet, respectively.
The possible reason is that to maintain a high prediction accuracy of the model, the percentage of patches dropped and shuffled of DBAVT cannot be set too high when defending against \sys. As a result, \sys is robust to DBAVT.
\begin{table*}[tt]
\vspace{0.1cm}
	\caption{Time cost of \sys and baseline attacks.}
	\label{tab:time}
	\centering
	\footnotesize
	\setlength\tabcolsep{7pt}
	\begin{tabular}{l|cccccccc}
		\toprule
		 Dataset&DBIA& BAVT &DBAVT-BadNets& TrojViT&DBAVT-WaNet & \sys \\
		\midrule
	    CIFAR-10&11.43h&6.35h&0.98h&1.74h&3.13h&1.67h\\
            GTSRB&3.23h&1.46h&0.25h&1.05h&0.67h&\color{black}0.62h\\
	    CIFAR-100&8.78h&4.52h&0.69h&1.67h&2.12h&1.65h\\
	    Tiny ImageNet&17.91h&10.42h&1.95h&1.04h &5.10h&3.76h\\
		\bottomrule
	\end{tabular}
\end{table*}

\textbf{BAVT.} BAVT \cite{subramanya2022backdoor} is based on the discovery that the interpretation map generated by transformers can effectively highlight the trigger used in a backdoored image. BAVT blocks out the region of the image with the highest values in the interpretation heatmap, thereby reducing the success rate of backdoor attacks.

We apply BAVT to \sys, and the defense results are shown in Table~\ref{tab:bavt}. We can see that \sys is robust to BAVT, as it maintains a high attack success rate even after BAVT is applied. This is likely due to the fact that BAVT uses the zero-setting operation on the region with the highest attention rollout in the test image, whereas \sys maximizes the attention rollout in the diffusion region rather than the poisoning region. Consequently, BAVT tends to cover non-poisoning areas within the diffusion area, making it susceptible to \sys.

\textbf{TeCo.} TeCo \cite{liu2023detecting} detects trigger samples at test time and requires only the victim models' hard label outputs. It capitalizes on the observation that backdoor-infected models behave similarly under various image corruptions with clean images, yet inconsistently with trigger samples. In the experiments, we adapted TeCo from CNNs to ViTs and then deployed it on poisoned samples from \sys. To determine an appropriate threshold, we initially selected 50 samples. However, the original method of calculating the threshold, which involved averaging the variance of recorded severity across all samples, significantly reduced the accuracy of identifying clean samples (dropping from 86.00\% to 45.2\% on the Tiny ImageNet dataset). Consequently, we opted to use the maximum variance as the threshold. As shown in Table~\ref{tab:teco}, we found that even after filtering out the backdoored samples with TeCo, \sys can also maintain a high attack success rate, reaching up to 75\%.

\textbf{Beatrix.} Beatrix \cite{ma2022beatrix} identifies poisoned samples by detecting anomalies in activation patterns. It utilizes Gram matrix to capture both feature correlations and high-order information of representations. While Beatrix effectively detects poisoned samples within CNNs, we discovered that it is not readily adaptable to ViT models, resulting in its failure to detect \sys.
In CNNs, different channels within the same layer represent distinct features from a single input image, making Gram matrix calculation useful for capturing interrelations among these diverse features. However, in ViT, each token only conveys information about a specific image patch. The Gram matrix computation in ViT lacks the capacity to encapsulate interconnections between varied representations across the entire image, making it challenging to distinguish between backdoor and benign samples.

\textbf{SAGE.} SAGE \cite{23spgong} is a newly-proposed backdoor purification method. It aims to correct toxic deep layers in a neural network by leveraging attention map alignment with innocent shallow layers. SAGE employs a layer-wise top-down self-attention distillation (SAD) technique to purify backdoored models. We first adapted SAGE from CNN models to vision transformers and then applied SAGE to the backdoored transformers trained with the poisoned samples from \sys.

The results are shown in Table~\ref{tab:sage}. It is shown that SAGE cannot effectively purify the backdoor from the backdoored models. \sys can also achieve ASR of 91.15\% (CIFAR-10), 78.29\% (GTSRB), 85.24\% (CIFAR-100), and 74.30\% (Tiny ImageNet) after being purified by SAGE defense.
\sys can successfully evade the SAGE backdoor defenses, possibly because that \sys has damaged the attention mechanism of different layers of the vision transformer rather than only poisoning the deeper layers as in CNN. Thus, our generated backdoor in the transformer cannot be purified by SAGE.

\section{User Study}
We conducted two sets of user studies to evaluate the concealment of poisoned samples of \sys. Fifty volunteers aged 20-30, including college students and faculty members, were recruited for the study. Prior to the user study, we provided a thorough explanation of \sys to the volunteers and confirmed their understanding of the attacks. For testing, we randomly selected 80 poisoned images from CIFAR-10, GTSRB, CIFAR-100, and Tiny ImageNet. 

\subsection{Similarity Test}
In the first test, we presented benign target samples and corresponding poisoned samples side-by-side to the volunteers. The benign target samples were displayed on the left, and the poisoned samples were displayed on the right. The volunteers were asked to judge the similarity between the two samples using a scale ranging from 1 to 5. A score of 5 represented ``look exactly the same", 4 represented ``very similar", 3 represented ``a little similar", 2 represented ``not very similar", and 1 represented ``very different". The cumulative distribution function (CDF) of the similarity scores is shown in Fig.~\ref{fig:user-study1}(a).  Experimental results demonstrate that over 50\% of the poisoned samples generated by \sys exhibit a similarity score surpassing 4.9, providing evidence that the poisoned samples are visually natural and capable of evading human visual inspections.
\begin{table}[tt]
	\caption{Apply DBAVT to \sys.}

	\label{tab:dbavt}
	\centering
	\footnotesize
	\setlength\tabcolsep{3pt}
	\begin{tabular}{l|cccccccc}
		\toprule
		 \multirow{2}{*}{\shortstack{Dataset}}&\multicolumn{3}{c}{Original} &\multicolumn{3}{c}{DBAVT}\\ &CDA& SCDA& SASR& CDA& SCDA& SASR \\
		\midrule
	    CIFAR-10&96.70\%&91.11\%&94.99\%&88.10\%&77.23\%&96.05\%\\
            GTSRB&96.27\%&93.20\%&91.00\%&94.05\%&84.14\%&81.34\%\\
	    CIFAR-100&87.26\%&95.13\%&97.00\%&80.36\%&77.26\%&100.0\%\\
	    Tiny ImageNet&88.20\%&90.06\%&90.02\%&80.40\%&68.01\%&94.89\%\\
		\bottomrule
	\end{tabular}
 
\end{table}

\begin{table}[tt]
	\caption{Apply BAVT to \sys.}
	\label{tab:bavt}
	\centering
	\footnotesize
        \setlength\tabcolsep{3pt}
	\begin{tabular}{l|cccccccc}
		\toprule
		\multirow{2}{*}{\shortstack{Dataset}}&\multicolumn{3}{c}{Original} &\multicolumn{3}{c}{DBAVT}\\ & CDA&SCDA& SASR & CDA&SCDA& SASR\\
		\midrule
	    CIFAR-10&96.70\%&91.11\%&94.99\%&96.70\%&90.99\%&75.00\%\\
            GTSRB&96.27\%&93.20\%&91.00\%&96.27\%&83.00\%&94.07\%\\
	    CIFAR-100&87.26\%&95.13\%&97.00\%&87.26\%&91.20\%&80.01\%\\
	    Tiny ImageNet&88.20\%&90.06\%&90.02\%&88.20\%&94.64\%&70.73\%\\
		\bottomrule
	\end{tabular}
\end{table}

\begin{table}[tt]
	\caption{Apply Teco to \sys.}
	\label{tab:teco}
	\centering
	\footnotesize
        \setlength\tabcolsep{3pt}
	\begin{tabular}{l|cccccccc}
		\toprule
		\multirow{2}{*}{\shortstack{Dataset}}&\multicolumn{3}{c}{Original} &\multicolumn{3}{c}{Teco}\\ & CDA&SCDA& SASR& CDA&SCDA& SASR\\
		\midrule
	    CIFAR-10&96.70\%&91.11\%&94.99\%&95.36\%&90.60\%&78.60\%\\
            GTSRB&96.27\%&93.20\%&91.00\%&94.99\%&93.00\%&92.30\%\\
	    CIFAR-100&87.26\%&95.13\%&97.00\%&83.23\%&94.99\%&76.51\%\\
	      Tiny ImageNet&88.20\%&90.06\%&90.02\%&86.00\%&88.98\%&75.25\%\\
		\bottomrule
	\end{tabular}
\end{table}
\begin{table}[tt]
	\caption{Apply SAGE to \sys.}
	\label{tab:sage}
	\centering
	\footnotesize
        \setlength\tabcolsep{3pt}
	\begin{tabular}{l|cccccccc}
		\toprule
		\multirow{2}{*}{\shortstack{Dataset}}&\multicolumn{3}{c}{Original} &\multicolumn{3}{c}{SAGE}\\ & CDA&SCDA& SASR& CDA&SCDA& SASR\\
		\midrule
	    CIFAR-10&96.70\%&91.11\%&94.99\%&97.00\%&91.00\%&91.15\%\\
            GTSRB&96.27\%&93.20\%&91.00\%&96.73\%&93.04\%&78.29\%\\
	    CIFAR-100&87.26\%&95.13\%&97.00\%&83.89\%&94.12\%&85.24\%\\
	      Tiny ImageNet&88.20\%&90.06\%&90.02\%&89.60\%&96.00\%&74.30\%\\
		\bottomrule
	\end{tabular}
\end{table}


\subsection{Inspection Test}
\begin{figure}[tt]
	\centering
	\begin{minipage}[t]{1.715in}
		\centering
		\includegraphics[trim=0mm 0mm 0mm 0mm, clip,width=1.715in]{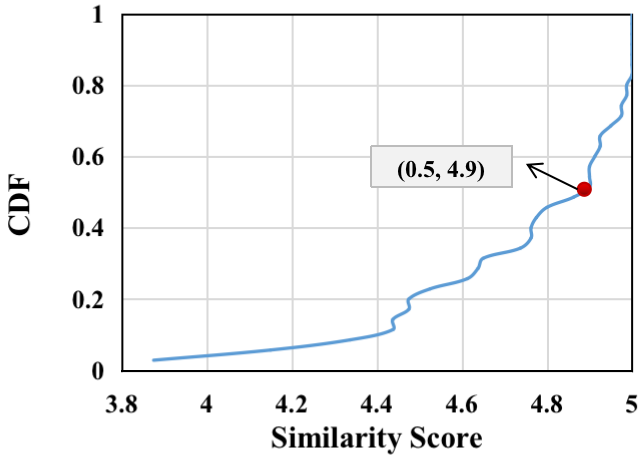}\\
		\centerline{\small (a) Similarity test}
	\end{minipage}
	\hspace{-0cm}
	\begin{minipage}[t]{1.68in}
		\centering
		\includegraphics[trim=0mm 0mm 0mm 0mm, clip,width=1.68in]{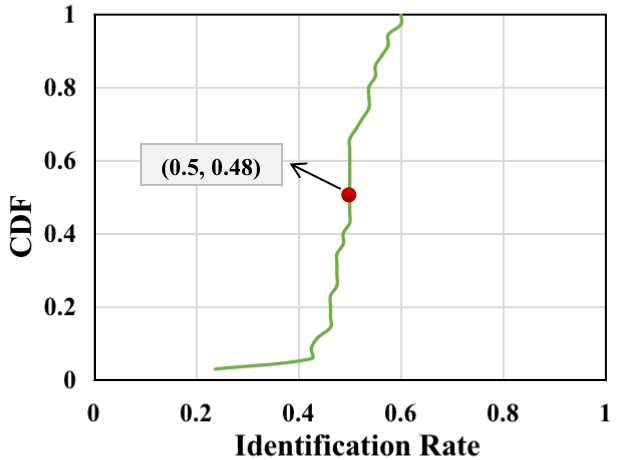}\\
		\centerline{\small (b) Inspection test}
	\end{minipage}
	\caption{{\color{black}User study results.} } \label{fig:user-study1}
\end{figure}

In the second test, we presented benign target samples and their corresponding poisoned samples side-by-side to the volunteers and asked them to choose which sample was the malicious one. To avoid any bias, we shuffled the position of the target and poisoned samples within each pair, so that the poisoned sample was not necessarily on the right. We calculated the percentage of correct answers as the identification rate. The cumulative distribution function (CDF) of the identification rate is shown in Fig.~\ref{fig:user-study1}(b). It is shown that approximately 50\% of the samples generated by \sys exhibit an identification rate lower than 0.48, akin to random guessing. This provides additional evidence supporting the evasiveness of \sys.

\section{Theoretical Analysis of Attention Diffusion Loss Effect}\label{8}
We start with $\mathbf{A}_{N}^{0,i}$, 
 representing the importance scores of the $i$-th token for the target label \cite{abnar2020quantifying}, and we simplified it as $\mathbf{A}^i$ for convenience. These scores reflect the contribution of the token towards the activation value $y_{tg}$ of the target label, without considering its impact on other labels ($y_{cl}$, where $cl \neq tg$). As the activation value of the target label is positively correlated with the probability of the model being classified as the target class, it also indicates the effectiveness of the backdoor attack.

Let $q$ represent the number of tokens in the attention diffusion area, $p$ denote the total number of tokens in the input sequence, and $m$ indicate the number of tokens in the poisoned area. In \sys, the value of $p$ is no less than $q$ to achieve the concealment goal.
When the attention diffusion area $q$ is equal to {\color{black}the poisoned area $m$ (patched trigger area)}, we refer to the attention diffusion loss as $\mathcal{L}_{\beta}$, which satisfies the following equations:
\begin{equation}
    \mathcal{L}_{\beta}= \sum_{i=1}^{m}(1-\mathbf{A}^i)+\sum_{i=m+1}^{p-1}\mathbf{A}^i.
\end{equation}
$\mathbf{A}^i$ has a positive contribution to the activation value of the poisoned area and a negative contribution to the non-poisoned area, as the coefficient in the poisoned area is 1 and -1 in the non-poisoned area. Therefore, $\mathcal{L}_{\beta}$ is negatively correlated with the effectiveness of \sys.
During the {\color{black}trigger generation process}, $\mathcal{L}_{\beta}$ is minimized, meaning it will be maximized within the diffusion range and minimized outside the diffusion range.

\begin{table*}[tt]
	\caption{Apply \sys to NLP}
	\label{tab:nlp}
	\centering
	\footnotesize
	\begin{tabular}{c|cc cc cc cc}
		\toprule
		\multirow{2}{*}{\shortstack{Tasks}}  &  \multicolumn{2}{c}{Original} &   \multicolumn{2}{c}{Original+$\mathcal{L}_{\alpha}$} 
      &     \multicolumn{2}{c}{Original+$\mathcal{L}_{\beta}$}   &     \multicolumn{2}{c}{Original+$\mathcal{L}_{\alpha}$+$\mathcal{L}_{\beta}$} \\ 
        &CDA&ASR     &    CDA&ASR   &   CDA&ASR    &   CDA&ASR  \\
		\midrule
	    Sentiment Analysis   & 91.68\% &91.85\% & 92.07\%  &  99.55\% &92.02\% & 99.33\% &91.46\%& 99.88\%  \\
            Toxic Detection  &95.17\% &79.52\% & 95.00\% &96.21\%  & 95.13\%&96.29\% & 94.72\% & 98.60\% \\
	    Topic Classification  & 94.28\%& 92.72\%& 94.46\% & 98.29\% & 94.04\%& 98.97\%& 94.19\%& 100\% \\
		\bottomrule
	\end{tabular}
\end{table*}

We evaluate two trigger generation losses, namely $\mathcal{L}_{\alpha}$ and $\mathcal{L}_{\alpha}+\mathcal{L}_{\beta}$. After generating, we denote $\mathbf{A}^i$ as $\mathbf{A}_{\Delta}^i$ when using $\mathcal{L}_{\beta}$, and as $\mathbf{A}_{0}^i$ when not using $\mathcal{L}_{\beta}$. Similarly, we denote the loss $\mathcal{L}_{\beta}$ as $\mathcal{L}_{\beta\Delta}$ when using $\mathcal{L}_{\beta}$, and as $\mathcal{L}_{\beta0}$ when not using $\mathcal{L}_{\beta}$, respectively.
\begin{equation}
\begin{split}
    \mathcal{L}_{\beta\Delta}=\sum_{i=1}^m(1-\mathbf{A}_{\Delta}^i)+\sum_{i=m+1}^{p-1}(\mathbf{A}_{\Delta}^i),\\
    \mathcal{L}_{\beta0}=\sum_{i=1}^m(1-\mathbf{A}_{0}^i)+\sum_{i=m+1}^{p-1}(\mathbf{A}_{0}^i),
\end{split}
\end{equation}
when $q \geq m$, 
\begin{equation}
\label{algo:1}
\begin{split}
    \mathcal{L}_{\beta\Delta}-
    \mathcal{L}_{\beta0}=(m-\sum_{i=1}^m\mathbf{A}_{\Delta}^i+\sum_{i=m+1}^{p-1} \mathbf{A}_{\Delta}^i)\\-(m-\sum_{i=1}^m \mathbf{A}_{0}^i+\sum_{i=m+1}^{p-1}\mathbf{A}_{0}^i)\\
    =[\sum_{i=1}^m(\mathbf{A}_{0}^i-\mathbf{A}_{\Delta}^i)-\sum_{i=q+1}^{p-1}(\mathbf{A}_{0}^i-\mathbf{A}_{\Delta}^i)]\\+
    (\sum_{i=m+1}^{q}\mathbf{A}_{\Delta}^i-\sum_{i=m+1}^{q}\mathbf{A}_{0}^i))\\
    < 0+ (\sum_{i=m+1}^{q}\mathbf{A}_{\Delta}^i-\sum_{i=m+1}^{q}\mathbf{A}_{0}^i)\\
    =\sum_{i=m+1}^{q}\mathbf{A}_{\Delta}^i-\sum_{i=m+1}^{q}\mathbf{A}_{0}^i,
\end{split}
\end{equation}
when $q=m$, the upper limit of of E.q.(\ref{algo:1}) is 0, i.e., $\mathcal{L}_{\beta\Delta}<\mathcal{L}_{\beta0}$. To this end, when the diffusion area is controlled within the poisoning area, the attention diffusion loss will enhance the {\color{black}attack effect}.

When $q>m$ and the attention diffusion area is not too large, although the upper bound of $\mathcal{L}_{\beta\Delta}-\mathcal{L}_{\beta0}$ is positive, $\mathcal{L}_{\beta\Delta}$ is generally smaller than $\mathcal{L}_{\beta0}$ due to the negative term in the square brackets.  However, the enhanced {\color{black}attack effect} will have a negative correlation with the diffusion area, thus if the attention diffusion area becomes too large, it will result in a decrease in the attack success rate. In \sys, we restrict the attention diffusion area to be less than three times the size of the trigger, which allows the attention diffusion loss to enhance the attack performance of \sys. 

\section{Discussion}

\subsection{Generalize \sys to NLP}
In this work, we design an evasive clean-label backdoor attack tailored for the vision transformer architecture. We show it may be possible to generalize \sys beyond the vision domain, extending it to the NLP domain.



To showcase \sys's adaptability and effectiveness in the NLP field, we select a state-of-the-art clean-label NLP backdoor attack framework \cite{lyu-etal-2023-attention} and replace its TAL loss with our diffusion loss, $\mathcal{L}{\beta}$, and latent loss, $\mathcal{L}{\alpha}$. We evaluate \sys across three diverse NLP task datasets, specifically Sentiment Analysis on the Stanford Sentiment Treebank (SST-2), Toxic Detection on HSOL, and Topic Classification on AG's News. As shown in Table~\ref{tab:nlp}, we can see that either $\mathcal{L}_{\beta}$ or $\mathcal{L}_{\alpha}$ can markedly enhance the attack performance. Notably, when both loss functions are applied, the attack success rate experiences a substantial uplift—from 91.85\% to 99.88\% (SST-2), from 79.52\% to 98.60\% (HSOL), and from 92.72\% to 100\% (AG's News). These findings unequivocally demonstrate that the loss functions in \sys can be effectively adapted to the NLP domain. Moving forward, we will delve deeper into crafting more sophisticated NLP-domain-specific backdoor attacks based on \sys, aiming to further refine attack performance.

\subsection{Potential Countermeasures}

Through experiments, we show that \sys is resistant to existing state-of-the-art defense strategies, including DBAVT, BAVT, Beatrix, TeCo, and SAGE. Design effective defenses against \sys are still required to mitigate the potential risks posed by such attacks.
One possible defense is analyzing the feature space of each collected training sample. The defender may conduct feature analysis on known benign data and collected data. If there is a distinct difference in feature space distribution, an alarm may be raised. However, this approach requires a large-enough known benign dataset to train a sufficiently accurate feature, otherwise, the false positive rate may be high, which leads to false rejection of benign data. To enhance defense, the defenders may further resort to creating an auto-encoder to detect the distribution of the difference in features. In the future, we will design effective defenses against \sys.

\section{Conclusion}

This paper presents an effective and evasive clean-label backdoor attack against vision transformers. We have carefully designed the trigger generation and trigger masking algorithms to achieve a high attack success rate with imperceptible triggers. Extensive experiments have demonstrated the superiority of our proposed attack compared to baseline backdoor attacks. There are several potential avenues for future research. Firstly, it may be possible to generalize \sys beyond the vision domain, extending it to other domains such as voice, text, and video. Secondly, there is a need for more sophisticated optimization in the algorithms for trigger generation and poisoned sample generation to further enhance attack performance, improve concealment, and reduce time costs. Last, effective defenses against \sys are necessary to mitigate the potential risks posed by such attacks.



%
\bibliographystyle{plain}
\bibliography{main}

\begin{IEEEbiography}[{\includegraphics[width=1in,height=1.25in,clip,keepaspectratio]{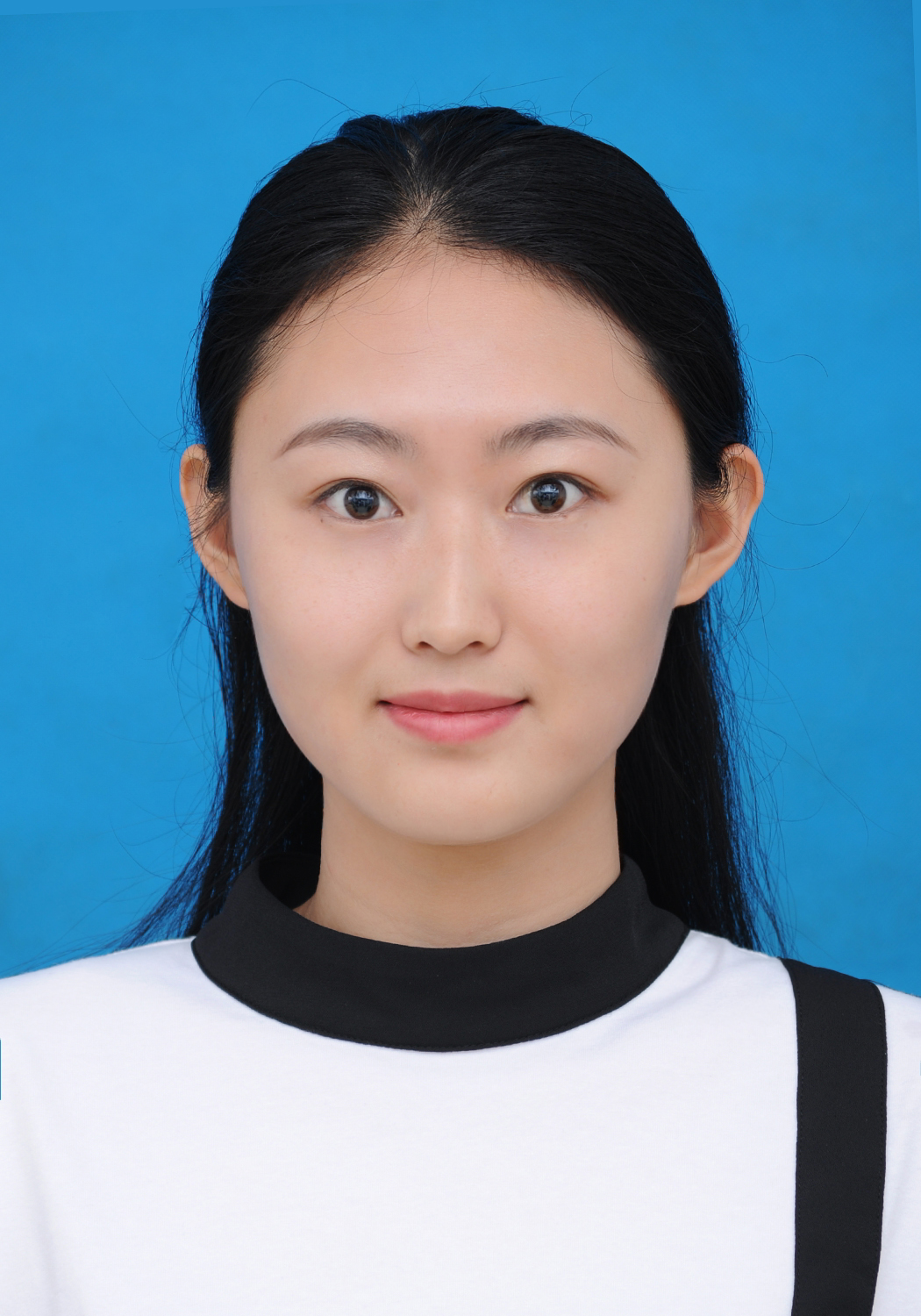}}]{Xueluan Gong} received her B.S. degree in Computer Science and Electronic Engineering from Hunan University in 2018. She received her Ph.D. degree in Computer Science from Wuhan University in 2023, China. Her research interests include network security, AI security, and data mining. She has published more than 20 publications in top-tier international journals or conferences, including IEEE S\&P, NDSS, Usenix Security, WWW, ACM Ubicomp, IJCAI, IEEE JSAC, TDSC, etc.
\end{IEEEbiography}

\begin{IEEEbiography}[{\includegraphics[width=1in,height=1.2in,clip,keepaspectratio]{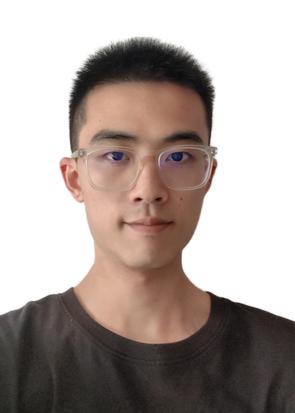}}]{Bowei Tian} is currently pursuing the bachelor degree in the School of Cyber Science and Engineering at Wuhan University, China. His research interests include information security and AI security.
\end{IEEEbiography}

\begin{IEEEbiography}[{\includegraphics[width=1in,height=1.2in,clip,keepaspectratio]{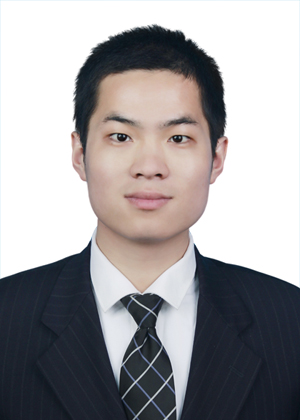}}]{Meng Xue} received his Ph.D. degree in Computer Science School from Wuhan University in 2022. He is currently a Postdoc in the Department of Computer Science and Engineering of Hong Kong University of Science and Technology. His research interests include Internet of things, smart sensing and and AI security.
\end{IEEEbiography}

\begin{IEEEbiography}[{\includegraphics[width=1in,height=1.2in,clip,keepaspectratio]{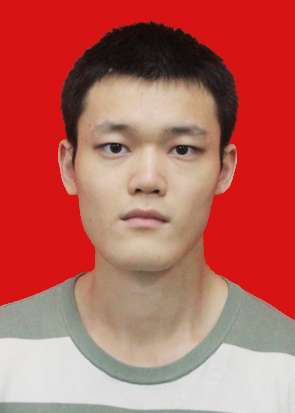}}]{Shuaike Li} is currently is currently pursuing the bachelor degree in the School of Cyber Science and Engineering at Wuhan University, China. His research interests include information security and AI security.
\end{IEEEbiography}

\begin{IEEEbiography}[{\includegraphics[width=1in,height=1.25in,clip,keepaspectratio]{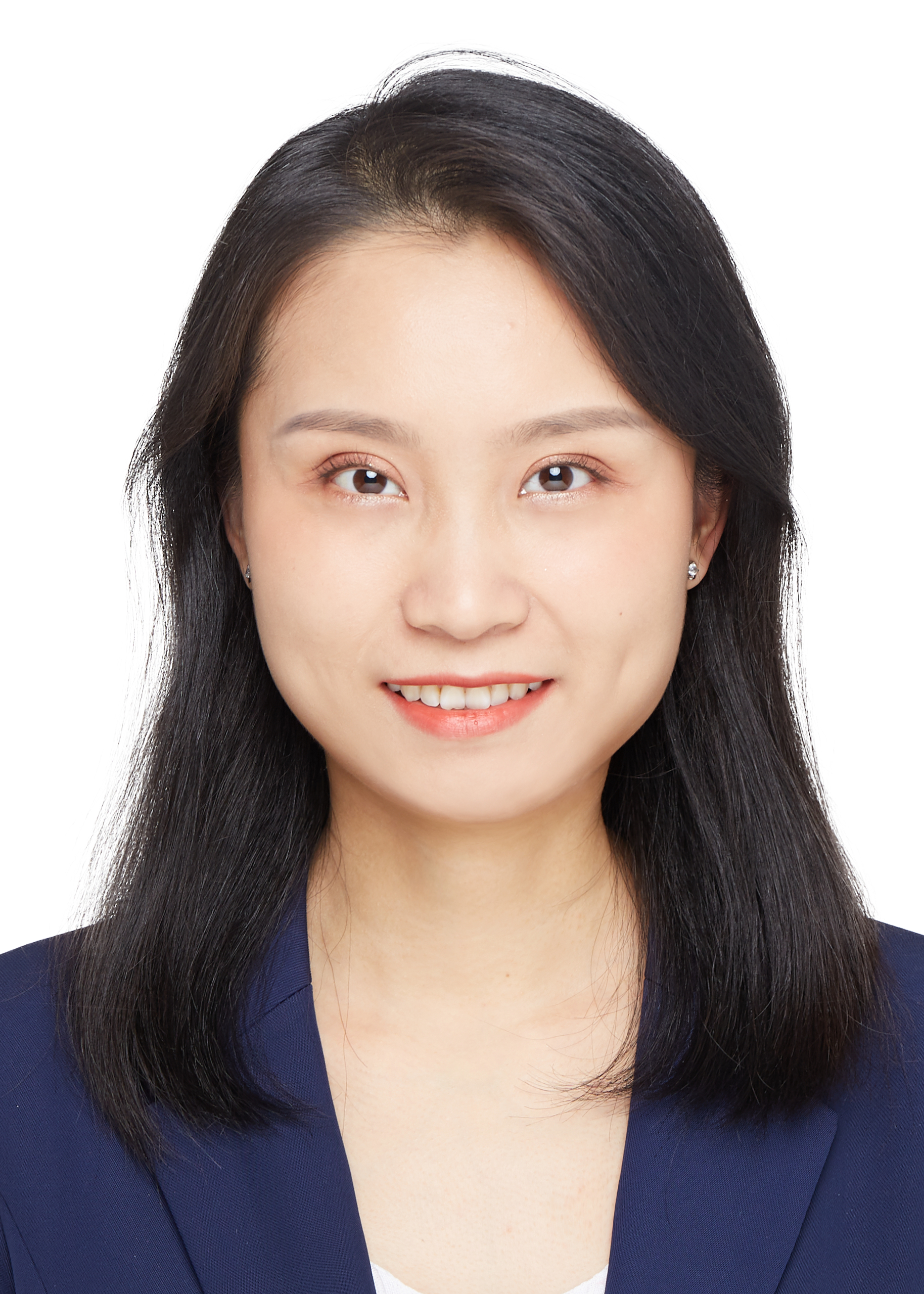}}]{Yanjiao Chen} received her B.E. degree in Electronic Engineering from Tsinghua University in 2010 and Ph.D. degree in Computer Science and Engineering from Hong Kong University of Science and Technology in 2015. She is currently a Bairen researcher in Zhejiang University, China. Her research interests include spectrum management for Femtocell networks, network economics, network security, AI security, and Quality of Experience (QoE) of multimedia delivery/distribution.
\end{IEEEbiography}

\begin{IEEEbiography}[{\includegraphics[width=1in,height=1.25in, clip,keepaspectratio]{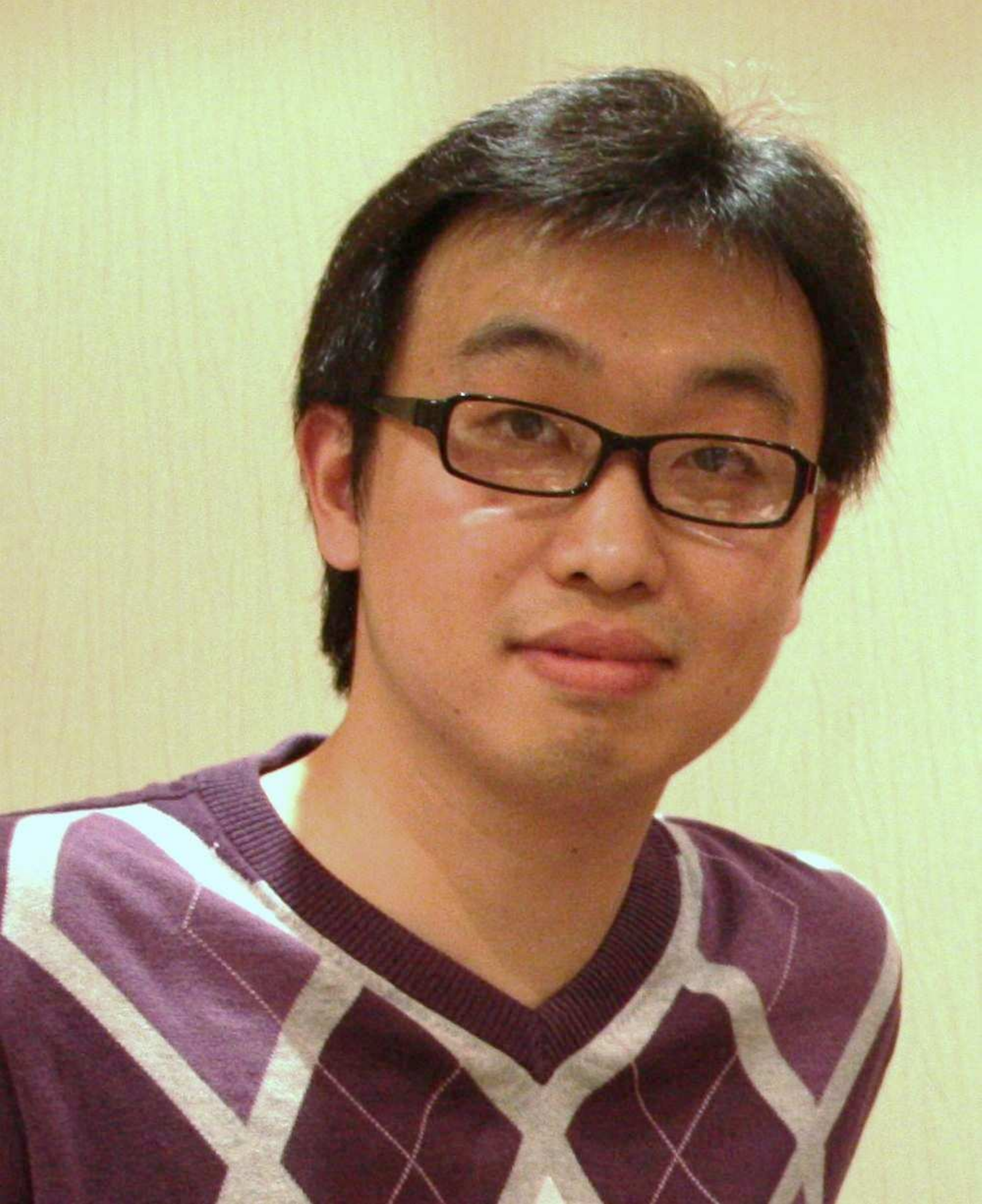}}]{Qian Wang} is a Professor in the School of Cyber Science and Engineering at Wuhan University, China. He was selected into the National High-level Young Talents Program of China, and listed among the World's Top 2\% Scientists by Stanford University. He also received the National Science Fund for Excellent Young Scholars of China in 2018. He has long been engaged in the research of cyberspace security, with focus on AI security, data outsourcing security and privacy, wireless systems security, and applied cryptography. He was a recipient of the 2018 IEEE TCSC Award for Excellence in Scalable Computing (early career researcher) and the 2016 IEEE ComSoc Asia-Pacific Outstanding Young Researcher Award. He has published 200+ papers, with 120+ publications in top-tier international conferences, including USENIX NSDI, ACM CCS, USENIX Security, NDSS, ACM MobiCom, ICML, etc., with 20000+ Google Scholar citations. He is also a co-recipient of 8 Best Paper and Best Student Paper Awards from prestigious conferences, including ICDCS, IEEE ICNP, etc. In 2021, his PhD student was selected under Huawei's  ``Top Minds'' Recruitment Program. He serves as Associate Editors for IEEE Transactions on Dependable and Secure Computing (TDSC) and IEEE Transactions on Information Forensics and Security (TIFS). He is a fellow of the IEEE, and a member of the ACM.
\end{IEEEbiography}

\end{document}